\documentclass{article}
\usepackage{ijcai23}

\usepackage{times}
\usepackage{soul}
\usepackage{url}
\usepackage[hidelinks]{hyperref}
\usepackage[utf8]{inputenc}
\usepackage[small]{caption}
\usepackage{graphicx}
\usepackage{amsmath}
\usepackage{amsthm}
\usepackage{booktabs}
\usepackage{algorithm}
\usepackage{algorithmic}
\usepackage[switch]{lineno}

\usepackage[para,online,flushleft]{threeparttable}
\usepackage{amssymb}
\usepackage{adjustbox}
\usepackage{color,xcolor}
\usepackage{epsfig}
\usepackage{subfigure}

\title{FedDWA: Personalized Federated Learning with Dynamic Weight Adjustment\footnote{An extended version of this paper (with the Appendix included) can be found in http://arxiv.org/abs/2305.06124.}}

\author{
    Jiahao Liu$^{1,2}$, \, Jiang Wu$^{1,2}$, \, Jinyu Chen$^{1,2}$, \, Miao Hu$^{1,2}$, \, Yipeng Zhou$^{3}$, \, Di Wu$^{1,2}$ \footnote{Corresponding author.}
    \affiliations
    $^{1}$School of Computer Science and Engineering, Sun Yat-sen University, Guangzhou, China
    \affiliations
    $^{2}$Guangdong Key Laboratory of Big Data Analysis and Processing, Guangzhou, China
    \affiliations
    $^{3}$School of Computing, Faculty of Science and Engineering, Macquarie University, Sydney, Australia
    \emails
    \{ liujh69, wujiang7, chenjy585  \}@mail2.sysu.edu.cn
    \emails
    \{ humiao5, wudi27\}@mail.sysu.edu.cn, yipeng.zhou@mq.edu.au
}

\begin{document}

\maketitle

\begin{abstract}
    Different from conventional federated learning, personalized federated learning (PFL) is able to train a customized model for each individual client according to its unique requirement.
The mainstream approach is to adopt a kind of weighted aggregation method to generate personalized models, in which weights are determined by the loss value or model parameters among different clients.
However, such kinds of methods require clients to download others' models. It not only sheer increases communication traffic but also potentially infringes data privacy. In this paper, we propose a new PFL algorithm called \emph{FedDWA (Federated Learning with Dynamic Weight Adjustment)} to address the above problem, which leverages the parameter server (PS) to compute personalized aggregation weights based on collected models from clients. In this way, FedDWA can capture similarities between clients with much less communication overhead. More specifically, we formulate the PFL problem as an optimization problem by minimizing the distance between personalized models and  guidance models, so as to  customize aggregation weights for each client. Guidance models are obtained by  the local one-step ahead adaptation on individual clients. Finally,  we conduct extensive experiments using five real datasets and the results demonstrate that FedDWA can significantly reduce the communication traffic and achieve much higher model accuracy than the state-of-the-art approaches.
\end{abstract}

\section{Introduction}
Federated Learning (FL), as an emerging distributed machine learning paradigm,  allows decentralized clients to collaboratively train a global machine learning model without exposing their private data \cite{mcmahan2017communication}.
However, one of the most challenging problems confronted by FL is the performance degradation caused by the heterogeneity of data distribution on decentralized clients \cite{Li2020OnTC}. 
Specifically, data distribution on clients is non-independent and identically distributed (non-IID) such that  a single global model cannot meet personalized needs of all clients. It has been reported in \cite{Li2020OnTC,Yu2020SalvagingFL} that data heterogeneity can result in slow convergence and poor model accuracy. For example, the global next-word prediction model trained by FedAvg \cite{mcmahan2017communication} is not always effective for all clients because of their personalized habits. Such a single global model may significantly deviate from personalized optimal models \cite{Yu2020SalvagingFL}. 

To address the above problem, \emph{Personalized Federated Learning (PFL)} has been proposed and studied in \cite{DBLP:conf/nips/SmithCST17,t2020personalized,fallah2020personalized,li2021ditto,collins2021exploiting}. PFL aims to handle non-IID data distribution by training personalized models for each client, so as to improve model accuracy. In essence,  PFL can either incorporate personalized components into the global model or train multiple models to obtain personalized models.
For example, the works \cite{t2020personalized,li2021ditto} 
added regularization terms to the global model in order to train personalized models. A distance metric  to shrink the search space of personalized models around the global model is applied. However, such an approach fails to optimize personalized models because   distance metrics usually cannot exactly capture the heterogeneity of data distribution among clients. Later on, more radical approaches (e.g., \cite{Zhang2021PersonalizedFL,li2022learning}) were proposed, which train multiple models to meet personalized requirements by distributing other clients' models to each individual client. Based on its local dataset, each client can decide how to aggregate models from other clients to obtain a personalized model. 
Despite that the performance of PFL is improved, this approach will make communication traffic explode and users' privacy may be compromised.

In this paper, we propose a novel PFL method called FedDWA (Federated Learning with Dynamic Weight Adjustment), which can improve PFL performance by encouraging collaborations among clients with similar data distributions. In existing works \cite{Zhang2021PersonalizedFL,li2022learning},  each client needs to collect models from all  other clients and evaluate similarities between clients with extra local validation sets. Different from these works, 
our framework  characterizes %and solve the coefficient estimation 
client similarity in an analytical way instead of empirical searching via the validation dataset. In addition, there is no need to share local models among clients in FedDWA avoiding excessive communication traffic and potential privacy leakage \cite{hu2021source}. In FedDWA, individual clients can obtain personalized models  computed by the PS based on collected model parameters and guidance models from clients. Guidance models are obtained using the one-step ahead adaptation method by individual clients. %, which indicate the local data distribution of clients.
Based on guidance models, the PS can %further train multiple personalized models for individual clients by 
tune aggregation weights to minimize the distance between each model with its guidance model. Thus, FedDWA can improve PFL performance without incurring heavy overhead by avoiding exchanging information between clients.

In summary, our main contributions in this paper can be summarized as follows:
\begin{itemize}
\item We propose a new personalized federated learning framework called FedDWA. 
FedDWA can effectively exploit clients owning data with a  similar distribution to improve personalized model accuracy.
% \textcolor{blue}{Compared with the latest relevant works, FedDWA not only achieves higher personalized model performance, but also reduces communication overhead largely.
% }
\item We theoretically analyze the properties of the FedDWA algorithm, and show how the weights are dynamically adjusted to achieve personalization. 
% FedDWA can effectively exploit clients with  similar data distribution to improve personalized model accuracy.
% \item Within this framework, we theoretically analyze why the design principles of the FedDWA algorithm can exploit clients with  similar data distributions to improve personalized model accuracy. \textcolor{red}{[What is the point of this sentence?]}
 \item By conducting experiments using five real datasets, we demonstrate that FedDWA outperforms other methods under three heterogeneous FL settings. 
\end{itemize}

\section{Related Work}
In this section, we discuss related works from two perspectives: \emph{data-based PFL} and \emph{model-based PFL}. Data-based PFL focuses on reducing data heterogeneity among clients while model-based PFL focuses on designing a personalized model for each client. Typically, data-based PFL shares a global dataset that is balanced across all clients \cite{zhao2018federated} (or private statistical information \cite{shin2020xor,yoon2021fedmix} among clients) to realize personalized learning. However, sharing a global dataset may potentially violate privacy policies since it is at the risk of privacy leakage. To address the above problem, model-based PFL was proposed, which can be divided into two types: \emph{single-model} PFL and \emph{multi-model} PFL. 

Most single-model PFL methods are extensions of conventional FL algorithms (e.g., FedAvg \cite{mcmahan2017communication}). For example, FedProx \cite{Sahu2020FederatedOI} employed a proximal term to formulate clients' optimization objectives so as to  mitigate the adverse influence of systematic and statistical heterogeneity on FL. FedAvg\_FT and FedProx\_FT \cite{wang2019federated} obtained personalized models by fine tuning the global model generated by FedAvg and FedProx, respectively. FedAvgM proposed by \cite{Hsu2019MeasuringTE} adopted a  momentum method to update the global model, so as to alleviate the adverse influence of non-IID data distribution on FL. An alternative single-model approach for PFL is  based on meta-learning.  Recent works \cite{DBLP:conf/nips/KhodakBT19,DBLP:conf/mobihoc/YueRXLZ21,acar2021debiasing} extended Model Agnostic Meta-learning (MAML) for FL under non-IID data distribution. 
However, the personalized learning ability of single-model methods is limited because it is hard to fit all heterogeneous data distributions with a single model very well. 

Multi-model methods outperform single-model methods by training multiple models to better adapt to the personalized requirements of clients. Cluster FL \cite{Sattler2021ClusteredFL,ghosh2020efficient,Mansour2020ThreeAF} assumed that clients can be partitioned into multiple clusters, and clients are grouped based on loss values or gradients. A customized model can be trained for each cluster. However, cluster-based client grouping may not be able to effectively improve PFL performance. 
FedEM \cite{marfoq2021federated} refined the cluster-based client group method by proposing a soft client clustering algorithm. However, it requires each client to download multiple models, which can considerably increase the communication overhead. 

Other than clustering clients, more advanced multi-model PFL methods were developed including additive model mixture between local and global models (such as L2GD \cite{hanzely2020federated} and APFL \cite{deng2020adaptive}),  multi-task learning methods with model similarity penalization (such as MOCHA \cite{DBLP:conf/nips/SmithCST17}, pFedMe \cite{t2020personalized} and Ditto \cite{li2021ditto}). 
More PFL methods were developed by leveraging Gaussian processes
\cite{achituve2021personalized} and knowledge transfer \cite{zhang2021parameterized}. However, these methods inevitably need public shared data or inducing points set. It is also possible to achieve PFL by decomposing FL models into a global part and multiple personalized parts. 
% The works \cite{DBLP:conf/nips/DinhTN20,hanzely2020federated,deng2020adaptive, dinh2021fedu,li2021ditto} trained more general models by considering simpler personalization terms to achieve a tradeoff between global and local models. 
Inspired by representation learning, the works \cite{arivazhagan2019federated,collins2021exploiting,tan2022fedproto,chen2021bridging,oh2021fedbabu,Mills2022MultiTaskFL} decomposed the FL model into a shared feature extractor part and a personalized part to realize PFL. Nevertheless, how to  decompose FL models is only heuristically designed and discussed by existing works. 

PFL can  be achieved by customizing weights of model aggregation for each client as well, \emph{e.g.},  
FedAMP \cite{huang2021personalized}, FedFomo \cite{Zhang2021PersonalizedFL} and L2C \cite{li2022learning}. These customizing weights  represent potential similarities between clients. \cite{ijcai2022p357} proposed that graph neural networks can also be used to learn  similarities among clients to realize personalization.
% optimized aggregation weights to generate personalized models in PFL.
FedAMP proposed an attentive message passing mechanism to compute personalized models, which is not flexible enough, because all clients need to participate in training in every  round.
FedFomo and L2C computed personalized aggregation weights via minimizing the validation loss on each client based on the model information collected from other clients, resulting in heavy communication traffic and concerns on privacy leakage. Although the effectiveness of customizing aggregation weights has been validated in existing works, their design is empirical based, not communication-efficient for large-scale real-world FL systems. Our work aligns with the line of work to customize aggregation weights in an analytical way without incurring heavy communication overhead.

\section{Problem Formulation}\label{Problem_Formulation}
In federated learning, each client $i$ owns a local private dataset denoted by $\mathcal{D}_{i}$ drawn from a distinct distribution $\mathcal{P}_{i}$. The objective of FL is to train a single global model $w$ for all clients by solving the following problem:
\begin{equation}
    \min _{w \in \mathbb{R}^{d}} \left\{ f(w):=\frac{1}{N} \sum_{i=1}^{N} f_{i}(w) \right\},
    \label{equation1}
\end{equation}
where the function $f_i: \mathbb{R}^{d}\to \mathbb{R}$ represents the expected loss over the data distribution of  client $i$, i.e., 
\begin{equation}
    f_{i}(w)=\mathbb{E}_{\xi_{i} \sim \mathcal{P}_{i}}\left[\tilde{f}_{i}\left(w ;\xi_{i}\right)\right].
\end{equation}
In the above equation, $\xi_i$ is a random sample generated according to the local distribution $\mathcal{P}_{i}$ and $\tilde{f}_{i}\left(w ;\xi_{i}\right)$ represents the loss function corresponding to sample $\xi_i$ and $w$. Since clients' data possibly come from different environments, they likely have non-IID data distributions, i.e., for $i \ne j$, $\mathcal{P}_{i} \ne \mathcal{P}_{j}$. 

In conventional FL, the target is to train a global model through conducting multiple global iterations. In the $(t-1)$-th global iteration, 
the PS distributes the latest global model parameters $w_{t-1}$ to all participating clients. The clients train locally and send the trained model $\hat{w}_{i}^{t}$ to PS for aggregation. The whole process can be shown as

\begin{equation}
    \hat{w}^{t}_{i} = w_{t-1} - \eta \nabla {f}_{i}(w_{t-1}), \text{ (training)}
\end{equation}
\begin{equation}
    w_{t} = \sum_{i=1}^{N}p_{i}\hat{w}_{i}^{t}.\qquad \qquad\text{(aggregation)}
    \label{eq:FedAvg_aggre}
\end{equation}
Here, we suppose that there are $N$ participating clients and 
$p_{i}$ is a pre-defined non-negative weight typically proportional to $|\mathcal{D}_i|$ with $\sum_{i=1}^{N}p_{i}=1$. $w_{t}$ represents the global model for the $t$-th round and $\eta$ is the learning rate.

%$w_{i}^{t}$ means the local model uploaded by client $i$ which has been trained locally at $t$ round, and 
From another perspective, the single global model trained by conventional FL is to minimize the L2 distance between a global model and all local models, which can be expressed as 
%a trade-off for all clients, because the global model $w_{t+1}$ can be thought the nearest center for all $\{w_{i}\}_{i=1}^{N}$ in terms of a weighted L2 distance:
\begin{equation}
    w_{t} = \mathop{\arg\min}_{w}\sum_{i=1}^{N}p_{i}\left\| w-\hat{w}_{i}^{t}\right\|^{2}.
    \label{solution1}
\end{equation}

However, if we consider the optimization of personalized models for individual clients, the optimization problem should be revised as
\begin{equation}
    \forall i, w_{i}^{\star} = \mathop{\arg\min}_{w_{i}}f_{i}(w_{i}).
    \label{PFL_problem}
\end{equation}
Here $w_{i}^{\star}$ represents the optimal target model for client $i$. 
If the data distribution is IID, it implies that  $ w_{i}^{\star} \approx   w_{i'}^{\star}$ for any two clients, which means that the optimal model applicable for each individual client can be derived by Eq.~\eqref{solution1}.
However, if the data distribution is non-IID, it has been investigated in \cite{li2019convergence,li2020federated} that a global model cannot satisfy all clients very well,  resulting in poor model accuracy. 

%For clients which have homogeneous data distribution, it has been shown that a single global model \cite{li2019convergence,li2020federated}. While for clients which have heterogeneous data distribution, as shown in Figure \ref{noniid_distribution}, suppose that the t communication round, the local model $w_{i}^{t}$ for client $i$ is close enough to the optimal model $w_{i}^{\star}$ after local training, then the new aggregated model $w_{t+1}$ may be worst than the local $w_{i}^{t}$ due to the discrepancy of clients' data distribution. To address the above limitation of a universal model, it is common to consider techniques that each client can learn its own specific personalized model. 

%In this setting, each client need to determine its own best model $w_{i}^{\star}$. 

% \begin{algorithm}[tb]
%     \caption{Example algorithm}
%     \label{alg:algorithm}
%     \textbf{Input}: Your algorithm's input\\
%     \textbf{Parameter}: Optional list of parameters\\
%     \textbf{Output}: Your algorithm's output
%     \begin{algorithmic}[1] %[1] enables line numbers
%         \STATE Let $t=0$.
%         \WHILE{condition}
%         \STATE Do some action.
%         \IF {conditional}
%         \STATE Perform task A.
%         \ELSE
%         \STATE Perform task B.
%         \ENDIF
%         \ENDWHILE
%         \STATE \textbf{return} solution
%     \end{algorithmic}
% \end{algorithm}

\begin{algorithm}[t]
\caption{FedDWA algorithm}
\label{alg:algorithm}
\textbf{Input}: Communication Round $T$, learning rate $\eta$, local epochs $E$, number of clients $N$, init model parameter $w^0$.
\textbf{Output}: Personalized model parameters $\mathbf{w}_1, \mathbf{w}_2, \dots, \mathbf{w}_N$.

\textbf{Server}
\begin{algorithmic}[1] %[1] enables line numbers
\FOR{$t=1,\dots, T$}
    \STATE  Server randomly selects a subset of clients  $S_t$ and sends $w_{1}^t,w_{2}^t,...,w_{m}^t$ to them.
    \FOR{each client $i \in S_t$ in parallel}
        \STATE $\hat{w}_{i}^{t}, \hat{w}_{i}^{\star} \gets$ Client($i$,$w_{i}^{t}$)
    \ENDFOR
    
    \STATE Compute $p_{i,j}$ according to Eq.~ \eqref{solution3} for each client $i$.
    % \For{each client $i\in S_t$}
    %     \For{$k \in S_{t}$}
    %          \State $p_{i,k}=\frac{\left\| \hat{w}_{i}^{\star}-w_{k}^{t}\right\|^{-2}}{\sum_{j=1}^{N}\left\| \hat{w}_{i}^{\star}-w_{j}^{t}\right\|^{-2}}$  
    %     \EndFor
    % \EndFor
    \FOR{each client $i\in S_t$}
        \STATE Select top-K clients $\{\mathcal{K}_{i}\}$.
        %\State $w_{i}^{t+1}=\sum_{k\in \mathcal{K}_{i}}p_{i k}w_{i}^{t}$
        \STATE Aggregate new model according to Eq.~\eqref{aggregation_step}.
    \ENDFOR
    
\ENDFOR
\end{algorithmic}

\textbf{Client}
\begin{algorithmic}[1]
\STATE $\mathcal{B}\gets$(split $\mathcal{D}$ into batches of size $B$)
\FOR{each local epoch $i$ from $1$ to $E$}
    \FOR{batch $ b \in \mathcal{B}$}
    \STATE$\mathbf{w} = \mathbf{w} -\eta \nabla \tilde{f}(\mathbf{w},b)$
    \ENDFOR
\ENDFOR
\STATE Train one more local iteration (one epoch).
\STATE $\hat{\mathbf{w}} = \mathbf{w} -\eta \nabla  \tilde{f}(\mathbf{w},\mathcal{D})$
\STATE \textbf{return} {$\mathbf{w}$ and $\hat{\mathbf{w}}$ } 
\end{algorithmic}
\end{algorithm}

\section{Methodology}
In this section, we elaborate the design of FedDWA and prove its effectiveness through analysis. 

\subsection{Optimization Objective}
In the previous section, it has been pointed out that the aggregation rule defined by Eq. \eqref{eq:FedAvg_aggre} cannot meet personalized requirement with non-IID data distribution. Rather than training a single global model, we propose to train a model for each individual client by customizing  aggregation weights so as to deduce each individual model. 

Specifically, we define $p_{i,j}$ as the weight to aggregate the model for client $i$ using the local model from client $j$ . Here,  $ \sum_{j=1}^{N} p_{i,j} = 1$. 
Then, the PS can generate the personalized model for client $i$ in the $(t-1)$-th global iteration as follows:
\begin{equation}
     \hat{w}^{t}_{i} = w_{i}^{t-1} - \eta_i^{t-1} \nabla {f}_{i}(w_{i}^{t-1}).
\end{equation}
\begin{equation}
    w_{i}^{t} = \sum_{j=1}^{N}p_{i,j}^{t}\hat{w}_{j}^{t}.
    \label{aggregation_step}
\end{equation}

Intuitively speaking, Eq.~\eqref{aggregation_step} is very flexible. When aggregating the model for client $i$, we can set a larger value for $p_{i,j}^{t}$ if the data distribution of client $j$ is closer to that of client $i$ such that the PS can explore optimal personalized model for client $i$\footnote{$p_{i,j}$ and $\eta_i$ can be time-dependent, but when context allows, we write $p_{i,j}^{t}$ as $p_{i,j}$ and $\eta_i^t$ as $\eta_i$ for simplicity.}. To specify how to set the value of $p_{i,j}$, we formulate the optimization problem as below: 

\begin{equation}
    \min_{p_{i,1},...p_{i,N}}\left\| \hat{w}_{i}^{\star}-\sum_{j=1}^{N}p_{i,j}\hat{w}_{j}^{t}\right\|^{2}, \quad \forall i.
    \label{problem2}
\end{equation}
Here $\hat{w}_{i}^{\star}$ is the guidance model for client $i$ in the $(t-1)$-th global iteration, and it tells client $i$ which  clients to cooperate with. The main challenge for solving Eq.~\eqref{problem2} lies in that $\hat{w}_{i}^{\star}$ is unknown in advance.  We can solve Eq.~\eqref{problem2} with two steps. Firstly, we need to find a high quality guidance model $\hat{w}_{i}^{\star}$ and fix $\hat{w}_{i}^{\star}$ to derive how to optimally set $p_{i,j}$. Secondly, after deriving the optimal weight $p_{i,j}$,  client $i$ can get its own personalized model at the  $t$-th round, and then its guidance model $\hat{w}_{i}^{\star}$ can be further updated. 
In the following, 
% we substitute $\hat{w}_{i}^{\star}$ for $w_{i}^{\star}$ and 
we elaborate how to solve Eq.~\eqref{problem2}. 

\subsubsection{Tuning Aggregation Weights}
Since $\sum_{j=1}^{N}p_{i,j}=1$, we can rewrite Eq.~\eqref{problem2} as: 
\begin{equation}
    \left\| \hat{w}_{i}^{\star}-\sum_{j=1}^{N}p_{i,j}\hat{w}_{j}^{t}\right\|^{2}\!\!\!\!=\sum_{j=1}^{N}\sum_{k=1}^{N}p_{i,j}p_{i,k}(\hat{w}_{i}^{\star}-\hat{w}_{j}^{t})^{T}(\hat{w}_{i}^{\star}-\hat{w}_{k}^{t}).
\end{equation}
Let the vector $\mathbf{p}_{i}=[p_{i,1}, p_{i,2}, ..., p_{i,N}]^{T}$ denote aggregation weights for obtaining client $i$'s personalized model. Let $\mathbf{W}_{i}$  denote the cross distance between the guidance model $\hat{w}^{\star}_{i}$  and local models contributed by clients. The $(j,k)$-th entry of $\mathbf{W}_{i}$ can be written as:
\begin{equation}
    [\mathbf{W}_{i}]_{j,k} = (\hat{w}_{i}^{\star}-\hat{w}_{j}^{t})^{T}(\hat{w}_{i}^{\star}-\hat{w}_{k}^{t}).
    \label{EQ: matrix_W}
\end{equation}
Then the optimization problem for client $i$ defined in Eq.~(\ref{problem2}) can be expressed  as follows:
\begin{equation}
\begin{aligned}
    &\min_{\mathbf{p}_{i}}\quad \mathbf{p}_{i}^{T}\mathbf{W}_{i}\mathbf{p}_{i}, \\
    &\text{subject to} \quad \mathbf{1}^{T}_{N} \mathbf{p}_{i}=1, p_{i,k} \ge 0.
\end{aligned}
\label{EQ:ConvertedOpt}
\end{equation}
Note that the PS can optimize personalized models for all clients via Eq.~\eqref{EQ:ConvertedOpt}.
Suppose that $\mathbf{W}_{i}$ is invertible, then the solution is given by:
\begin{equation}
\begin{aligned}
    \mathbf{p}_{i} = \frac{\mathbf{W}_{i}^{-1} \mathbf{1}_N}{\mathbf{1}_N^{\mathrm{T}} \mathbf{W}_{i}^{-1} \mathbf{1}_N}
\end{aligned}
\label{EQ:ClosedSolution}
\end{equation}
It is very difficult to directly calculate Eq.~(\ref{EQ:ClosedSolution}) due to the following three challenges. \textit{First}, it involves the inner product of  model parameters among clients which can be seen from Eq.~\eqref{EQ: matrix_W}. For advanced  neural networks, there may exist millions of parameters making the computation cost unaffordable. \textit{Second}, training models in federated learning is  an iterative process with multiple rounds of communications. It implies that computing the inversion of $\mathbf{W}_{i}$ is  cumbersome, especially when the dimension of $\mathbf{W}_{i}$ is very large. \textit{Third}, since $\mathbf{W}_{i}$ is just a symmetric matrix, $\mathbf{W}_{i}^{-1}$ may not exist at all. As a consequence,  the solution of Eq.~(\ref{EQ:ConvertedOpt}) is not unique. 
% For example, consider the case where $\mathbf{W}_{i}^{-1}$ has a rank less than $N-1$, we can find an infinite number of solutions. 
Thus, trying to solve Eq.~\eqref{EQ:ConvertedOpt} directly cannot guarantee  that a high-quality  solution will be yielded.
A toy example is discussed in Appendix \ref{appendix:tony_example}.

%In short, solving Eq.~(\ref{EQ:ConvertedOpt}) is difficult, whether or not $\mathbf{W}_{i}$ is invertible. 
To make the problem tractable, we simplify the objective in Eq.~(\ref{EQ:ConvertedOpt}) by  only reserving the diagonal elements of $\mathbf{W}_{i}$.
{The effectiveness of such simplification has been verified in 
previous works  \cite{chen2015diffusion,zhao2012clustering}.
}
The simplified problem is presented as follows:
\begin{equation}
\begin{aligned}
    &\min_{\mathbf{p}_{i}}\quad \sum_{j=1}^{N}p_{i,j}^2\left\| \hat{w}_{i}^{\star}-\hat{w}_{j}^{t}\right\|^{2}, \\
    &\text{subject to} \quad \mathbf{1}^{T}_{N} \mathbf{p}_{i}=1, p_{i,j} \ge 0.
\end{aligned}
\label{solution2}
\end{equation}
It is easy to find  that there is a unique solution to the simplified problem. It is worth noting that Eq.~\eqref{solution2} is very alike to the  aggregation rule in conventional FL defined in Eq.~\eqref{solution1}. The difference of our method can be explained from two perspectives. First, aggregation weights are tunable parameters in our method, which however are fixed in traditional FL. Second, our method can search the optimal aggregation weights for individual clients to derive personalized models. The solution of Eq.~\eqref{solution2} is:
\begin{equation}
    p_{i,j}=\frac{\left\| \hat{w}_{i}^{\star}-\hat{w}_{j}^{t}\right\|^{-2}}{\sum_{k=1}^{N}\left\| \hat{w}_{i}^{\star}-\hat{w}_{k}^{t}\right\|^{-2}},
    \label{solution3}
\end{equation}
where the detailed derivation can be found in Appendix \ref{appendix:solution_p}.

\subsubsection{One-step Ahead Adaptation}
% Now let us consider how to set the guidance model $w_{i}^{\star}$. 
% A naive idea is to set $\hat{{w}}_{i}^{t} = {w}_{i}^{\star}$, unfortunately, it is impossible to derive ${w}_{i}^{\star}$ since client $i$ has only a limited number of samples and cannot calculate the expected risk. Intuitively, whether we can achieve good personalization here involves figuring out which models benefit client $i$, so the weighted combination of models should best align with client $i$'s interests, which means that the guidance model should reflect the local data distribution of client $i$. Therefore, we can approximate $\hat{{w}}_{i}^{t}$ with the local one-step approximation method. 
Executing the combination rule in Eq. \eqref{solution3} by an individual client  requires the knowledge of the guidance model $\hat{w}_{i}^{\star}$, which is generally not available beforehand or not.
% since a single client does have the capability to compute the expected risk with  a limited number of samples. 
Intuitively speaking, whether we can realize personalization depends on similar clients identified by our algorithm. % involves figuring out which models benefit client $i$, so 
In other words, the weighted combination of models for client $i$ should align with client $i$'s personal data distribution. It implies that the guidance model $\hat{w}_{i}^{\star}$ should capture the local data distribution of client $i$.  We have tried several options for $\hat{w}_{i}^{\star}$ including using the last iteration model $\hat{w}_{i}^{\star} = \hat{w}_{i}^{t-1}$, current model $\hat{w}_{i}^{\star} = w_{i}^{t}$, and one-step ahead adaptation. More discussion can be found in Appendix \ref{appendix:selection_of_guidance_model}. In this work, we employ an instantaneous adaptation argument, a.k.a. local one-step ahead adaptation,  to accommodate this issue as follows:
% The validity of this approximation can be found in previous work \cite{jin2020affine,chen2015diffusion,zhao2012clustering}. Suppose that client $i$  gets $w_{i}^{t}$ by conducting $E$ local iterations with its local data set, then client $i$ derives  $\hat{w}_{i}^{t}$ with one more local iteration as below:
\begin{equation}
    \hat{w}_{i}^{\star} = \hat{w}_{i}^{t}-\eta_{i}^{t-1}\nabla f_{i}(\hat{w}_{i}^{t}).
    \label{one-step_approximation}
\end{equation}

The validity of this adaptation can be found in previous work \cite{jin2020affine,chen2015diffusion}. It is important to note that this is fundamentally different from traditional FedAvg training and then local fine-tuning, because our method will only select other clients that are beneficial to client $i$ for aggregation while the fine-tuning approach treats all clients equally during the aggregation phase. In fact, after a step of adaptation, Eq.~\eqref{one-step_approximation} in advance, $\hat{w}_{i}^{\star}$ can characterize its local data distribution well and therefore it can screen out other clients with similar data distribution and give them a higher weight for cooperation. We can also use two steps or even more steps, the specific experimental results will be shown in the following sections. 
By substituting Eq. \eqref{one-step_approximation} into Eq.~\eqref{solution3}, we can finally derive the aggregation weights by deriving personalized models as
\begin{equation}
    p_{i,j}=\frac{\left\| \hat{w}_{i}^{t}-\eta_{i}^{t-1}\nabla f_{i}(\hat{w}_{i}^{t})-\hat{w}_{j}^{t}\right\|^{-2}}{\sum_{k=1}^{N}\left\| \hat{w}_{i}^{t}-\eta_{i}^{t-1}\nabla f_{i}(\hat{w}_{i}^{t})-\hat{w}_{k}^{t}\right\|^{-2}}.
    \label{solution4}
\end{equation}
Similar to the previous works \cite{Zhang2021PersonalizedFL,li2022learning}, the top-K technique can be used. We rank $p_{i,j}$ in descending order, and only the top $K$ aggregation weights are selected and they will be normalized such that $\sum_{j=1}^{N}p_{i,j}=1$. By wrapping up our analysis, we present the detailed FedDWA algorithm in Algorithm.~\ref{alg:algorithm}.

\begin{table*}[t]
\begin{tabular}{l|ccccc|cccc}
\hline
Settings & \multicolumn{5}{c|}{Pathological heterogeneous setting} & \multicolumn{4}{c}{Practical heterogeneous setting 1} \\ \hline
Methods & EMNIST & CIFAR10 & CIFAR100 & CINIC10 & TINY & CIFAR10 & CIFAR100 & CINIC10 & TINY \\ \hline
Local Training & 97.23 & 92.35 & 80.08 & 92.74 & 58.04 & 72.12 & 39.82 & 64.40 & 19.90 \\ \hline
FedAvg & 71.78 & 59.97 & 34.71 & 46.07 & 9.64 & 71.57 & 44.67 & 58.97 & 13.74 \\
FedProx & 69.95 & 57.42 & 31.12 & 45.60 & 8.46 & 70.03 & 41.53 & 56.01 & 6.40 \\
FedAvgM & 66.42 & 59.82 & 34.00 & 48.81 & 8.84 & 71.49 & 44.56 & 58.77 & 12.22 \\
FedAvg\_FT & 93.65 & 88.82 & 62.73 & 81.87 & 18.30 & 74.35 & 46.76 & 62.45 & 14.70 \\ 
\hline
SFL & 72.02 & 60.71 & 34.17 & 48.83 & 10.88 & 71.77 & 44.83 & 59.28 & 13.46 \\
per-FedAvg & 93.45 & 88.44 & 62.83 & 89.65 & 16.28 & 73.32 & 43.62 & 62.57 & 8.70 \\
pFedMe & 95.42 & 90.78 & 78.10 & 88.72 & 51.12 & 77.68 & 45.79 & 59.82 & 17.26 \\
ClusterFL & 78.45 & 83.41 & 51.30 & 80.48 & 40.26 & 71.42 & 44.90 & 59.44 & 12.92 \\
FedRoD & 93.77 & 87.95 & 64.76 & 83.75 & 50.96 & 77.27 & 46.76 & 52.64 & 18.04 \\ \hline
FedAMP & 96.65 & 91.74 & 78.61 & 88.05 & 50.72 & 72.30 & 41.11 & 65.43 & 27.48 \\
FedFomo & 96.95 & 91.95 & 78.89 & 92.59 & 59.24 & 75.69 & 47.06 & 68.93 & 19.16 \\
L2C & 95.75 & 91.76 & 78.62 & 92.69 & 54.24 & 76.67 & 48.30 & 67.52 & 21.04 \\
 \hline
Ours & \bf{97.37} & \bf{92.97} & \bf{80.41} & \bf{92.75} & \bf{60.64} & \bf{78.09} & \bf{50.83} & \bf{70.29} & \bf{28.92} \\ \hline
\end{tabular}
\caption{Average test accuracy (\%) over five different datasets, under pathological heterogeneous setting and pracitical heterogeneous setting 1 with 20 clients, $100\%$ participation, respectively.}
\label{tab1}
\end{table*}

\subsection{Analysis of FedDWA}
% Next, we analyze the properties of FedDWA algorithm to explain the underlying reason for the efficiency of FedDWA to achieve personalized learning.

\subsubsection{Communication Overhead} %As shown in Figure \ref{fig2}, since clients can send multiple models in one upload to server, we still maintain one round of communication for model uploads and one
% Based on the presented FedDWA algorithm, we can analyze its communication cost. 
% % Compared with the original  FedAvg algorithm \cite{mcmahan2017communication}, there is no additional cost to distribute personalized models from the PS to individual clients. 
% For uplink, each client needs to upload both local models and the guidance model achieved by one-step approximation. Thus, the uplink communication traffic per global iteration is at most   $2P_{b}$, 
% \textcolor{blue}{and the downlink communication traffic is the same as the original FedAvg\cite{mcmahan2017communication}} where $P_{b}$ denotes the model size. 
The FedDWA algorithm will incur $2\Sigma$ traffic in the uplink communication and the traffic in the downlink communication is the same as that of the original FedAvg, where $\Sigma$ denotes the model size.  It is worth noting that FedDWA can incur significantly less communication traffic than other similar baselines, and more results can be found in Appendix \ref{appendix:communication_cost}.
% For example, the FedFomo algorithm  proposed in  \cite{Zhang2021PersonalizedFL} and the L2C algorithm proposed in \cite{li2022learning} need to download $K\Sigma$ and $N\Sigma$ communication traffic, respectively, in each global iteration to train personalized models. Here, $K$ is the hyperparameter of top $K$ strategy, meaning the number of collaborating clients in FedFomo and $N$ is the total number of  clients in L2C. Although these algorithms only need to upload $\Sigma$ communication traffic, in general, our method can significantly reduce the amount of communication traffic compared to FedFomo and L2C during each communication round, 
% which will be explored in next section.
% \textcolor{blue}{
% Here, $K$ is the hyperparameter of top $K$ strategy, meaning the number of collaborating clients in FedFomo and $N$ is the total number of  clients in L2C. Although these algorithms only need to upload $\Sigma$, in general, our method can significantly reduce the amount of communication traffic compared to FedFomo and L2C during each communication round, which can be seen in Table \ref{tab3}.
% }

\subsubsection{Computational Cost}
Suppose that there are $N$ clients participating in training and the number of model parameters is $d$, the computation complexity of FedDWA is $ \mathcal{O}(N^2d)$ in the server. We also test the total FLOPs required by FedDWA and the experimental results show that the computational amount required to calculate the similarity (Eq.(15)) is negligible compared with model training. More results can be found in Appendix  \ref{appendix:computational_cost}.

\subsubsection{Personalized Learning}
In this part, we illustrate how FedDWA can make clients with similar data distribution collaborate to  train  personalized models. Considering the inverse of the numerator of Eq.~\eqref{solution4}, it can be expanded as
\begin{equation}
\begin{aligned}
    &\left\| \hat{w}_{i}^{t}-\eta_{i}^{t-1}\nabla f_{i}(\hat{w}_{i}^{t})-\hat{w}_{j}^{t}\right\|^{2}=\\ &\left\| \hat{w}_{i}^{t}-\hat{w}_{j}^{t}\right\|^{2}+2\eta_{i}^{t-1}(\hat{w}_{j}^{t}-\hat{w}_{i}^{t})^{T}\nabla f_{i}(\hat{w}_{i}^{t})+\\ &(\eta_{i}^{t-1})^2 \left\|\nabla f_{i}(\hat{w}_{i}^{t})\right\|^{2}.
\end{aligned}
\end{equation}
The first term $\left\| \hat{w}_{i}^{t}-\hat{w}_{j}^{t}\right\|^{2}$ refers to the distance   between current models of client $i$ and client $j$.
This term will lower  the aggregation weight $p_{i,k}$ if the distance $\left\| \hat{w}_{i}^{t}-\hat{w}_{j}^{t}\right\|^{2}$  is large, and thereby prohibit their collaborations.  
Using the first-order Taylor series to expand $f_{i}(w)$ at $\hat{w}_{i}^{t}$, we have:
\begin{equation}
    f_{i}(w)\approx f_{i}(\hat{w}_{i}^{t}) + (w-\hat{w}_{i}^{t})^{T}\nabla f_{i}(w)\mid_{\hat{w}_{i}^{t}}.
\end{equation}
For the second term $2\eta_{i}^{t-1}(\hat{w}_{j}^{t}-\hat{w}_{i}^{t})^{T}\nabla f_{i}(\hat{w}_{i}^{t})$,
we can find that it is proportional to $f_{i}(\hat{w}_{j}^{t})-f_{i}(\hat{w}_{i}^{t})$ which also decreases the aggregation weight $p_{i,k}$ if $f_{i}(\hat{w}_{j}^{t})$ is far away from $f_{i}(\hat{w}_{i}^{t})$. The last term $(\eta_{i}^{t-1})^2 \left\|\nabla f_{i}(\hat{w}_{i}^{t})\right\|^{2}$ can be perceived as a constant when optimizing aggregation weights. In summary, Eq. \eqref{solution4} provides the aggregation weights for deriving a personalized model for client $i$ based on the similarity distance between client models $\hat{w}_{j}$'s and the guidance model $\hat{w_{i}}^{\star}$ .

\section{Experiments}
\subsection{Experiment Setups}
\subsubsection{Datasets and Models}
We evaluate our algorithm on five benchmark datasets, namely, EMNIST \cite{cohen2017emnist}, CIFAR10, CIFAR100 \cite{krizhevsky2009learning}, CINIC10 \cite{darlow2018cinic} and Tiny-ImageNet (TINY) \cite{chrabaszcz2017downsampled}. 
% For EMNIST, we use a convolution neural network (CNN) with two convolution layers and two fully connected layers as that in \cite{Sattler2021ClusteredFL}. 
For EMNIST, we use the same model as that used in \cite{Sattler2021ClusteredFL}. 
% For CIFAR10, CIFAR100 and CINIC10, we use the CNN model which has two convolutional pooling layers, two batch normalization layers and two fully connected layers, as the same in \cite{Mills2022MultiTaskFL}.
For CIFAR10, CIFAR100 and CINIC10, we use the CNN model which is the same as that in \cite{Mills2022MultiTaskFL}.
% which has two convolutional pooling layers, two batch normalization layers and two fully connected layers, as the same in \cite{Mills2022MultiTaskFL}.
To evaluate the effectiveness of FedDWA on a high-dimensional model, we use ResNet-8 for the Tiny-ImageNet, and the model architecture is the same as that in  \cite{he2020group}. More details can be found in Appendix \ref{appendix: datasets_models}.

\subsubsection{Data Partitioning} We simulate the heterogeneous settings with three widely used scenarios, including a pathological setting and two practical settings. 
\begin{itemize}

    \item {\bf  Pathological Heterogeneous Setting.}  Each client is randomly assigned with a small number of classes with the same amount of data on each class \cite{mcmahan2017communication,shamsian2021personalized}. We sample 4, 2, 2, 6 and 10 classes for EMNIST, CIFAR10, CINIC10, CIFAR100, Tiny-ImageNet from a total of 62, 10, 10, 100, 200 classes for each client, respectively. There is no group-wise similarity between clients in this setting.
    \item  {\bf Practical Heterogeneous Setting 1.}  All clients have the same data size but different distributions. For each client, $s\%$ of data ($80\%$ by default) are selected from a set of dominant classes, and the remaining $(100-s)\%$ are uniformly sampled from all classes \cite{karimireddy2020scaffold,huang2021personalized}. All clients are divided into multiple groups. Clients in each group share the same dominant classes implying that there is an underlying clustering structure between clients.
    \item {\bf Practical Heterogeneous Setting 2.}   Each client contains most of the classes but the data in each class is not uniformly distributed \cite{Hsu2019MeasuringTE,li2021model,chen2021bridging}. We create the federated version by randomly partitioning datasets among $N$ clients using a symmetric Dirichlet distribution $\text{Dir}(\alpha)$ ($\alpha=0.07$ by default). For example, for each class $c$,  we sample a vector $p_{c}$ from $\text{Dir}(\alpha)$ and allocate to client $m$ a  fraction $p_{c,m}$ of all training instances of class $c$.
\end{itemize}

\subsubsection{Baselines} We compare the performance of our algorithm with that of  FedAvg \cite{mcmahan2017communication}, FedAvgM \cite{Hsu2019MeasuringTE}, FedProx \cite{li2020federated} and a few latest personalization approaches including a personalized model trained only on each client's local dataset (Local Training), FedAvg with local tuning (FedAvg\_FT) \cite{wang2019federated}, pFedMe \cite{t2020personalized}, per-FedAvg \cite{fallah2020personalized}, ClusterFL \cite{Sattler2021ClusteredFL}, FedAMP \cite{huang2021personalized}, FedFomo \cite{Zhang2021PersonalizedFL}, SFL\cite{ijcai2022p357}, L2C \cite{li2022learning} and FedRoD \cite{chen2021bridging}. The settings of hyper-parameters for each method can be found in Appendix \ref{appendix: implementation_of_baselines}.

\subsubsection{Evaluation Metrics} We use the same evaluation metrics as that widely used by previous works which report the test accuracy of the best single global model for the single-model PFL methods and the average test accuracy of the best personalized models for other PFL methods. 
% The higher the average test accuracy is, the better the method performance is. 

\subsubsection{Training Settings} 
%We implement our method and other baseline methods in PyTorch 1.7, and the simulation server is equipped with an Tesla V100 GPU, a 2.4-GHz Inter Core E5-2680 CPU and 256GB of memory. 
Similar to \cite{Zhang2021PersonalizedFL}, we evaluate the performance in two settings, i.e., (1) $N=20$ clients, $100\%$ participation and (2) $N=100$ clients, $20\%$ participation for all datasets. The number of local training epochs is set to $E=1$ and the number of global communication rounds is set to 100. %where all methods have little or no accuracy gain with more communications. 
We employ the mini-batch SGD as a local optimizer for all approaches. The batch size for each client is set as $20$ and the learning rate $\eta$ is set as $0.01$. We test all methods over three runs and average the results. 

\subsection{Performance Evaluation and Analysis}
\subsubsection{Pathological Heterogeneous Setting} Table \ref{tab1} shows the average test accuracy for all methods under the pathological heterogeneous setting. Over all datasets and client setups, our FedDWA algorithm outperforms other baseline methods.
%especially in the settings with a larger number of clients, and also has excellent performance in the  scenario with 20 clients. The reason is that when there are a larger number of clients, each individual client carries less local training data, and they have strong desire to exchange information with those who have similar data distribution through the federated updates. 
However, the performance of methods (such as FedAvg and FedProx) that only train a single global model degrade significantly on CIFAR10, CIFAR100, CINIC10 and Tiny-ImageNet, since a single  global model cannot well accommodate statistical heterogeneity of clients. 
Other personalized methods  achieve comparable accuracy and the local training method achieves rather high performance due to the small number of classes on each client. 

\begin{table}[t]
\scalebox{0.94}{
\begin{tabular}{lcccc}
\hline
 & CIFAR10 & CIFAR100 & CINIC10 & TINY \\ \hline
FedAvg & 48.44 & 22.80 & 26.68 & 6.72 \\
FedProx & 45.08 & 23.69 & 27.06 & 5.90 \\
FedAvgM & 48.09 & 21.70 & 27.36 & 4.70 \\
FedAvg\_FT & 89.19 & 45.06 & 83.32 & 6.47 \\ \hline
SFL & 52.40 & 22.96 & 36.82 & 5.13 \\
per-FedAvg & 89.86 & 49.81 & 89.69 &  18.52\\
pFedMe & 90.54 & 51.70 & 89.41 &  6.15\\
ClusterFL & 85.48 & 31.33 & 76.73 &  21.85\\
FedRoD & 86.83 & 42.90 & 88.72 &  26.77\\ \hline
FedAMP & 91.27 & 51.16 & 90.51 &  27.14\\
FedFomo & 91.73 & 54.29 & 91.32 &  32.33\\
L2C & 91.77 & 54.98 & 91.67 &  30.41\\ \hline
Ours & \bf{91.81} & \bf{55.26} & \bf{91.80} & \bf{32.66} \\ \hline
\end{tabular}
}
\caption{Average test accuracy (\%) over four different datasets, under the practical heterogeneous setting 2 with 100 clients, $20\%$ participation. }
\label{tab2}
\end{table}

\subsubsection{Practical Heterogeneous Setting} Table \ref{tab1} shows the average test accuracy for all methods under the practical heterogeneous setting 1 in which each client has a  primary data class with a small number of samples from all other classes. In this setting, we find that our method is significantly superior to all other baseline methods. For example, when using Tiny-ImageNet, our method outperforms FedFomo and L2C by up to $9.76\%$, $7.88\%$ in test accuracy respectively, which means that the aggregation weights obtained by our method are better than those with the aggregation weights  obtained  by empirical searching through the validation set. To test the applicability of our algorithm, we also evaluate the performance of our approach for a large-scale FL scenario under the practical heterogeneous setting 2. We set $N=100$ clients with $20\%$ participation in each round. The final results are shown in Table \ref{tab2}. It is worth noting that our approach still achieves competitive performance  though there is no clear similarity between clients in this scenario.

\subsubsection{Personalized Model Weighting}
\begin{figure}[t]
\centering
\subfigure[FedDWA]{
    % \label{fig:iter_l} 
    \includegraphics[width=0.99\columnwidth]{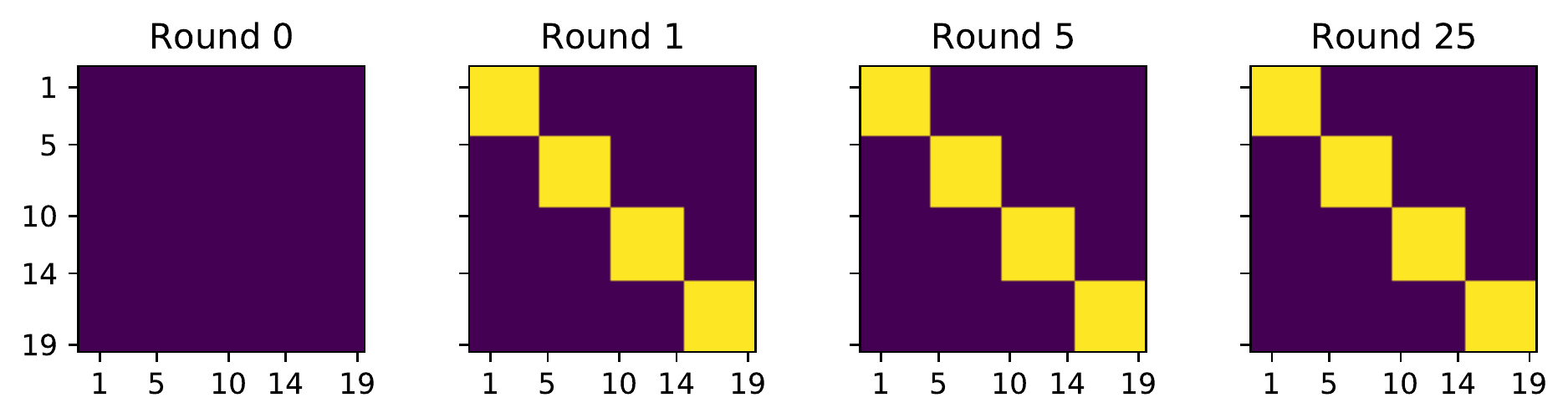}% Reduce the figure size so that it is slightly narrower than the column. Don't use precise values for figure width.This setup will avoid overfull boxes.
\label{FedDWA_weight}
}

\subfigure[FedFomo]{
    \includegraphics[width=0.99\columnwidth]{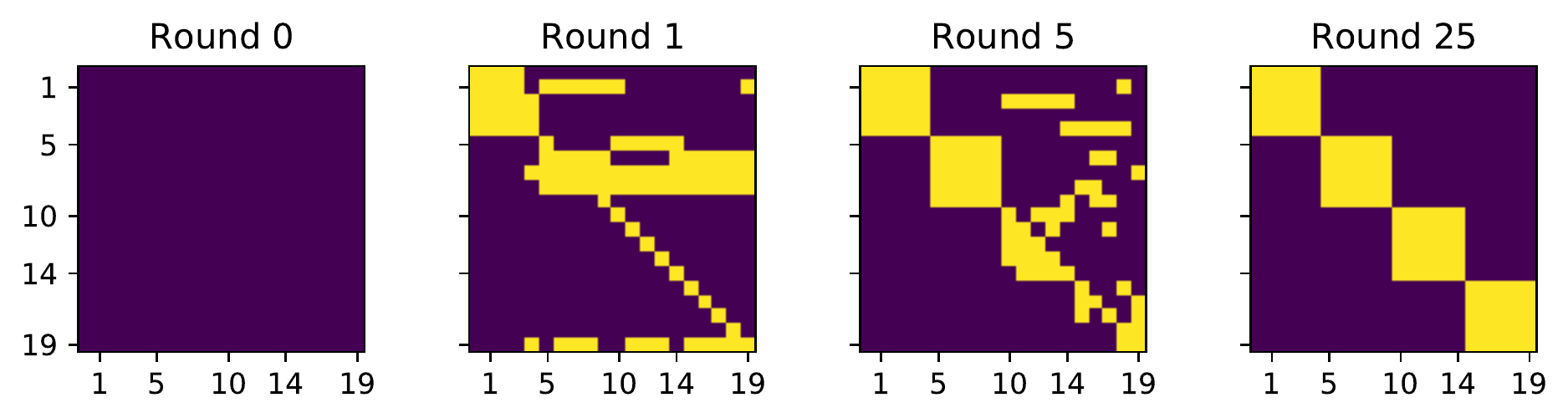}% Reduce the figure size so that it is slightly narrower than the column. Don't use precise values for figure width.This setup will avoid overfull boxes.
\label{FedFomo_weight}
}

\subfigure[L2C]{
    \includegraphics[width=0.99\columnwidth]{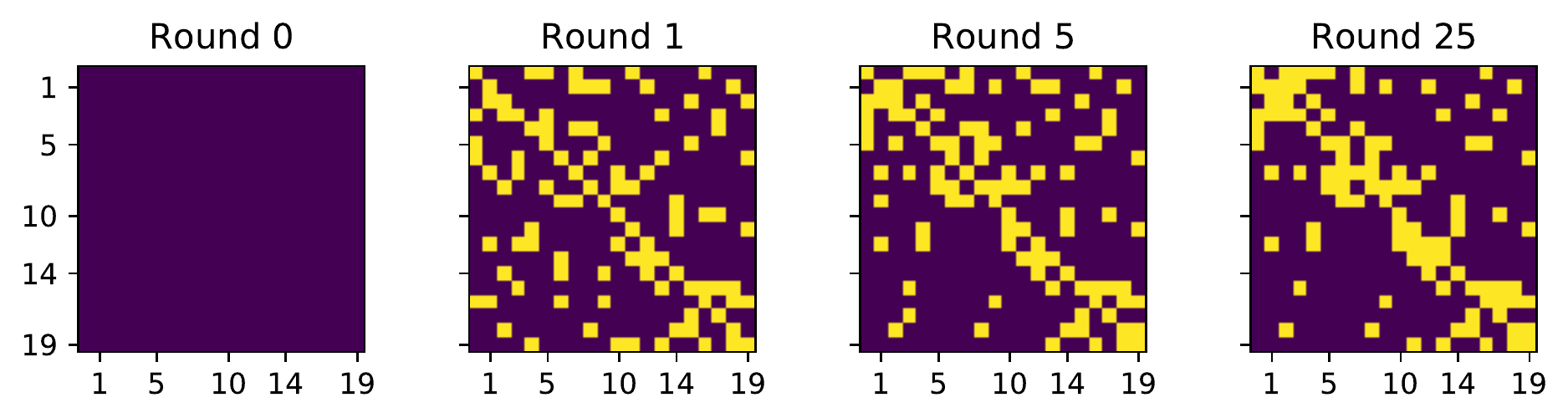}% Reduce the figure size so that it is slightly narrower than the column. Don't use precise values for figure width.This setup will avoid overfull boxes.
\label{L2C_weight}
}

\subfigure[FedAMP]{
    \includegraphics[width=0.99\columnwidth]{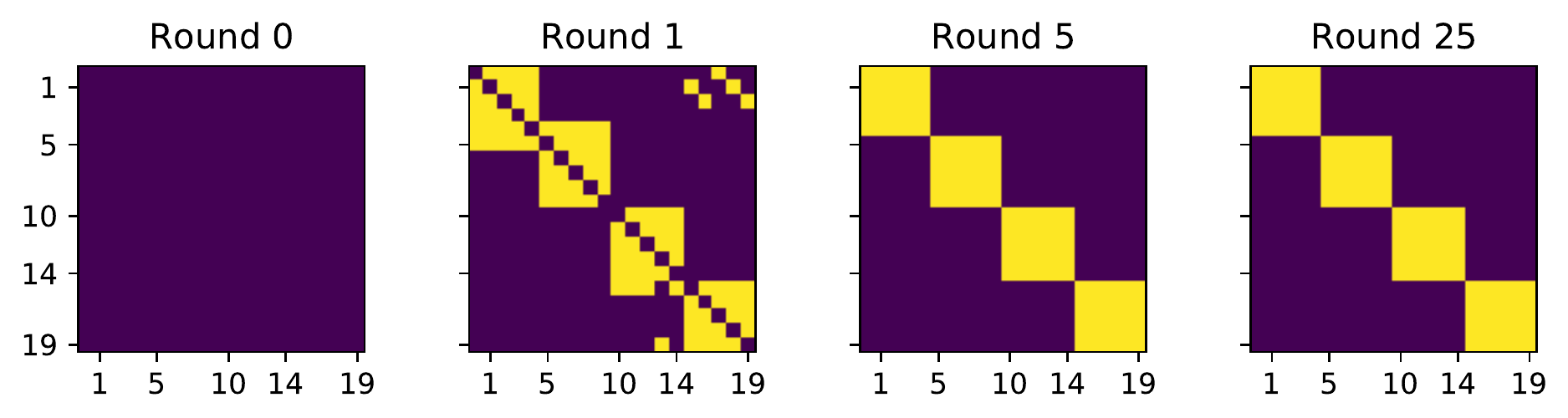}% Reduce the figure size so that it is slightly narrower than the column. Don't use precise values for figure width.This setup will avoid overfull boxes.
\label{FedAMP_weight}
}
\caption{The visualization of the $K$ most similar clients selected by different methods on CIFAR10 datasets. The x-axis and y-axis means the IDs of clients.}
\label{similarity}
\end{figure}

% In order to show how FedDWA allows clients to find their optimal personalized model by selecting other suitable clients, we next visualize the aggregation weights $p_{i,k}$ computed by FedDWA in Figure \ref{similarity}. 
We use the practical heterogeneous setting 1 in which clients are divided into multiple groups to explain why FedDWA outperforms existing works by showing how FedDWA helps clients quickly identify other similar clients for conducting personalized model aggregation.  
%The data distribution of clients in the same group is similar, while the data distribution of clients in different groups is discrepant. 
% Specifically, for clients in the same group,  $80\%$ of data samples of every client are uniformly sampled from a set of dominating classes, and $20\%$ of data samples are uniformly sampled from the rest of classes. 
% Therefore, each client has data from all classes, but the number of samples for each class is different. 
To ease our visualization, we group clients first before clients are indexed such that clients in the same group will have consecutive indices. 
%Specifically, We depict clients with the same local data distributions next to each other (
For example, clients of the same data distribution are grouped in the first group who are indexed by 0-4. %  that have similar data distribution and client 5-9 are in the same group). 
In Figure \ref{similarity}, we show the $K$ most similar clients selected by different methods according to personalized weights in each training round with 20 clients divided into 4 groups. Since there are five clients in each group, we set $K=5$. The result manifests that FedDWA can identify similar clients faster than FedFomo and L2C. In addition, under this setting, L2C cannot accurately identify the similarity between users, indicating that  weights found with validation datasets are not very effective. More experiment results such as the heat maps showing the aggregation weights can be found in Appendix \ref{appendix:additional_personalized_weights}.

\subsubsection{Selection of K}
\begin{figure}[t]
\centering
\includegraphics[width=0.7\columnwidth]{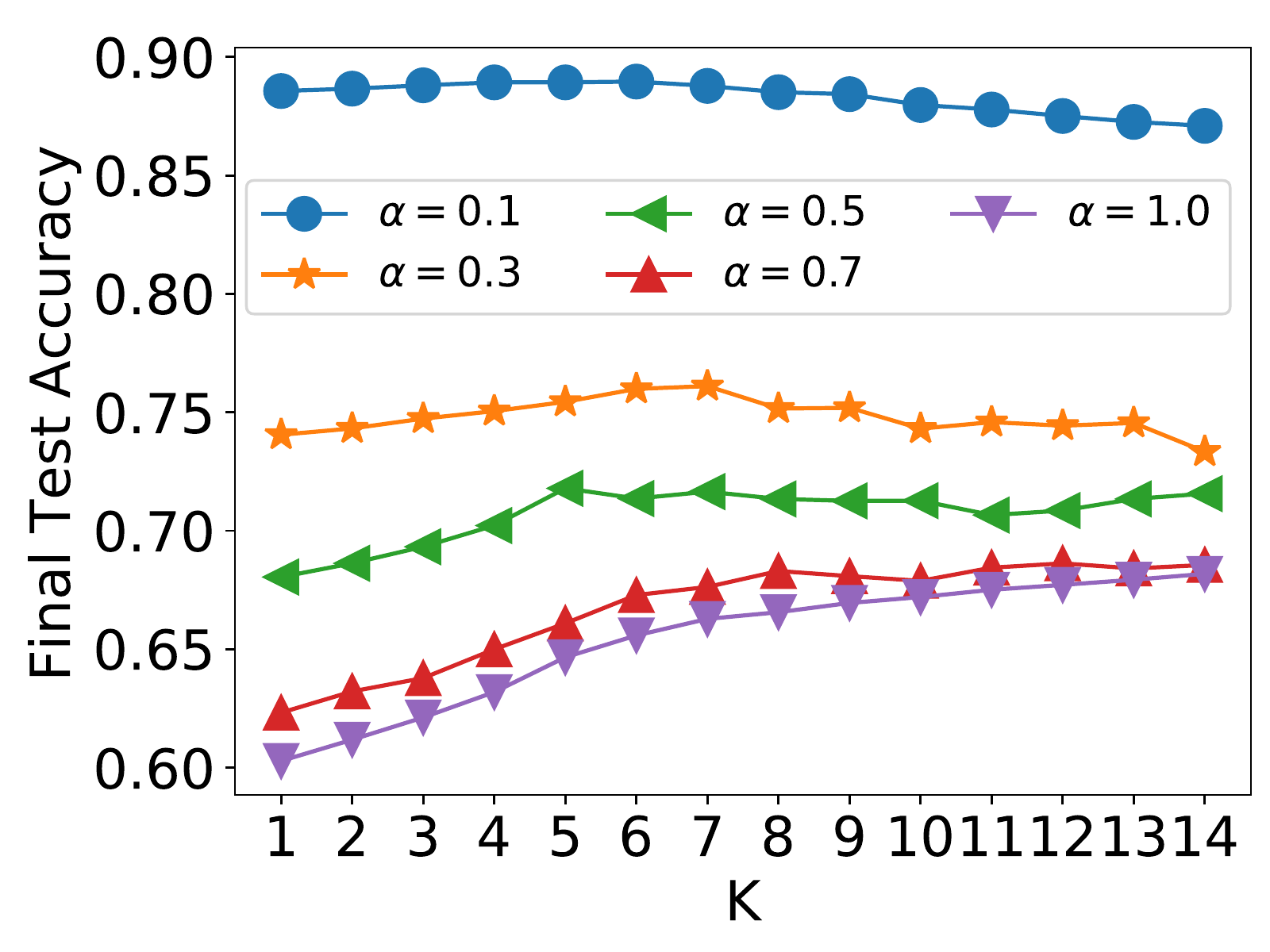}% Reduce the figure size so that it is slightly narrower than the column. Don't use precise values for figure width.This setup will avoid overfull boxes.
\caption{The influence of different values of K on the final performance of our algorithm using CIFAR10 dataset.}
\label{effect_k}
\end{figure}

Figure \ref{appendix: personalized_weights} in Appendix \ref{appendix:additional_experiments} shows that when there is a potential cluster structure among clients participating in FL, the weight calculated by FedDWA at the first round can reflect the similarity between clients well. At this time, server can easily determine the value of $K$ according to the weight matrix. However, as reported  in \cite{ghosh2020efficient,Zhang2021PersonalizedFL,li2022learning,marfoq2022personalized},  it is not trivial to set $K$ if there is no  obvious data similarity between clients.
Fortunately, the performance of FedDWA is not very sensitive with respect to $K$.  %sensitivity of FedDWA to the value of $K$ in a broad FL scenario, 
We use the practical heterogeneous setting 2 to simulate different degrees of data heterogeneity with 100 clients, and present the test performance in Figure \ref{effect_k}. 
% Note that there is no obvious similarity between clients in this setup. 
The results indicate that FedDWA is not very sensitive to the value of $K$ when data distributions are more heterogeneous (with a smaller $\alpha$). When $\alpha$ is large, the discrepancy between users' data distributions is smaller, which is not a typical PFL scenario. At this time it becomes more important to increase the value of $K$ so as to cooperate with more clients. %Besides, we notice that our method is not sensitive to the value of $K$ in some interval.
% While for CIFAR-10 and CIFAR-100, when the value of K exceeds a certain threshold, the performance of FedDWA decreases. As we have analysed in Eq.~\eqref{solution4}, it is likely that overfitting occurs when $N$ is too large, so it is essential to select a proper value of $K$.

\subsubsection{Effect of Guidance Model} 
\begin{figure}[t]
\centering
\includegraphics[width=0.7\columnwidth]{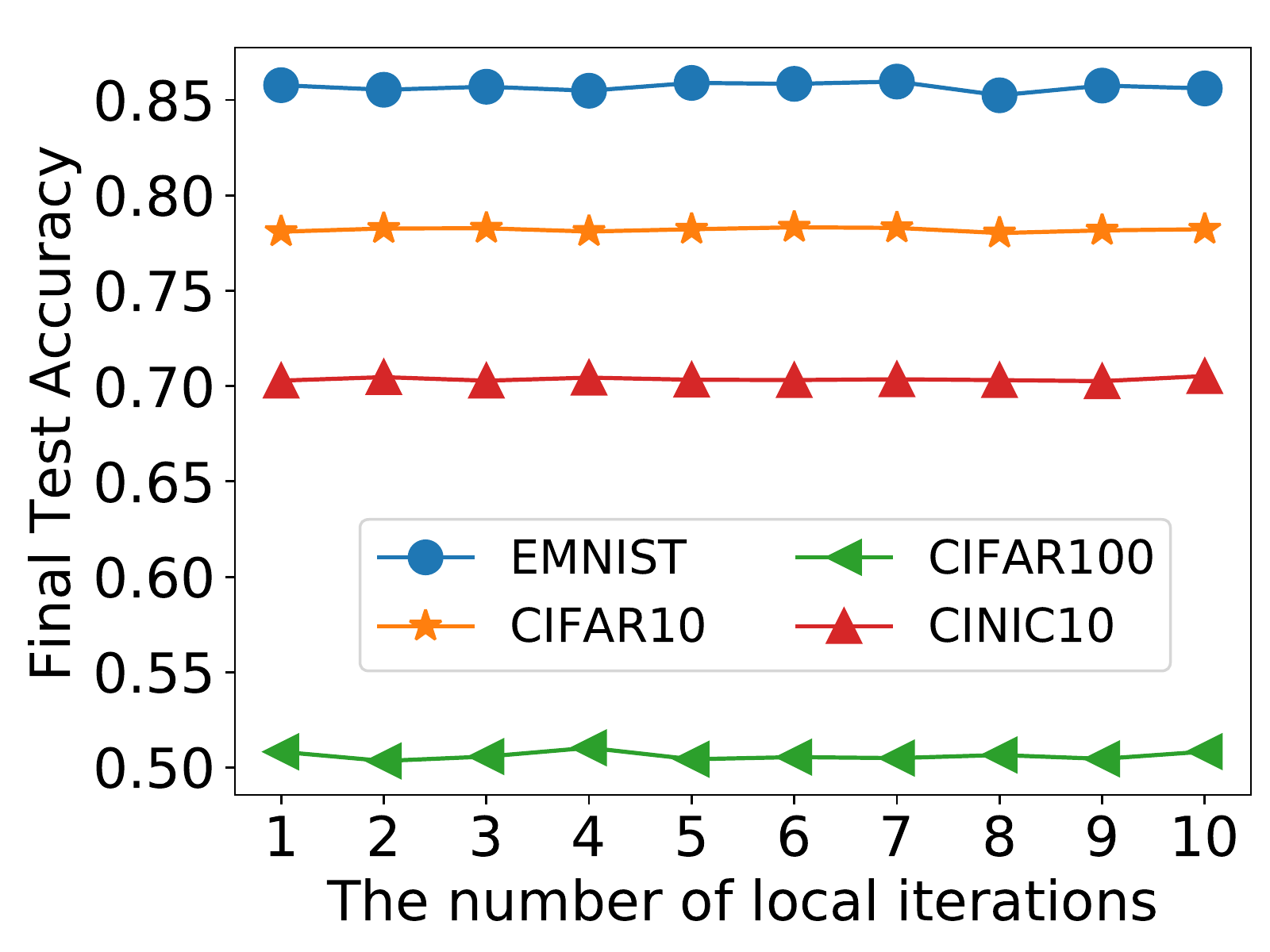}% Reduce the figure size so that it is slightly narrower than the column. Don't use precise values for figure width.This setup will avoid overfull boxes.
\caption{The influence of different local epoch iterations on the final test accuracy of our algorithm.}
\label{effect_targetModel}
\end{figure}

We explore the influence of guidance models obtained by Eq.~\eqref{one-step_approximation} on the final personalized model accuracy by tuning the number of local epochs. According to  Eq.~\eqref{one-step_approximation}, $\hat{w}_{i}^{\star}$ will be different if we set a larger number of local epochs.  We evaluate this effect with  20 clients on four datasets, respectively. %hrough different steps (one step is equivalent to one epoch) adaptation on FedDWA. 
As shown in Figure \ref{effect_targetModel}, the number of local iterations to compute guidance models has little influence  on the final model accuracy  of FedDWA indicating that one-step adaptation is sufficient for FedDWA. Moreover, we compare the performance of guidance model $\hat{w}_{i}^{\star}$ and personalized model $w_{i}^{t}$ in Appendix \ref{appendix:comparison_results}. The results further show that the performance of $w_{i}^{t}$ is almost the same as that of $\hat{w}_{i}^{\star}$ under the practical setting 2 and better than that of $\hat{w}_{i}^{\star}$ under the practical setting 1.

\section{Conclusions}
% Due to the  inherent non-IID data distribution in FL, it is essential and imperative to personalize FL models. How to effectively personalize FL without incurring heavy communication overhead is a challenging  problem because there is lack of effective approaches for a client to identify others with similar data distribution for collaborative training in FL.  
In this paper, we propose a novel FedDWA algorithm which can identify similarities between clients with much less communication overhead than other relevant works since no information will be exchanged between clients. Meanwhile, personalized models are generated based on uploaded model information and 
its effectiveness is guaranteed with theoretical analysis. 
%, which enables clients to approach their own optimal personalized model adapting to their local data distribution and achieves strong performance across two popular non-IID settings, especially when network traffic is limited. We formulate the entire training process where clients search for their personalized models as a least mean squares optimization problem and provide an approximated solution theoretically. Besides, since FedDWA only requires additional computation on the server, our method can therefore reduce the communication bandwidth compared with FedFomo. 
Comprehensive experiments on five datasets demonstrate the superb performance of FedDWA which can achieve the highest model accuracy compared to the state-of-the-art baselines under three heterogeneous FL settings.

\section*{Acknowledgements}
This work was supported by the National Natural Science Foundation of China under Grants U1911201, U2001209, 62072486, and the Natural Science Foundation of Guangdong Province under Grant 2021A1515011369.

%% The file named.bst is a bibliography style file for BibTeX 0.99c
\bibliographystyle{named}
\bibliography{ijcai23}

\begin{thebibliography}{}

\bibitem[\protect\citeauthoryear{Acar \bgroup \em et al.\egroup
  }{2021}]{acar2021debiasing}
Durmus Alp~Emre Acar, Yue Zhao, Ruizhao Zhu, Ramon Matas, Matthew Mattina, Paul
  Whatmough, and Venkatesh Saligrama.
\newblock Debiasing model updates for improving personalized federated
  training.
\newblock In {\em International Conference on Machine Learning}, pages 21--31.
  PMLR, 2021.

\bibitem[\protect\citeauthoryear{Achituve \bgroup \em et al.\egroup
  }{2021}]{achituve2021personalized}
Idan Achituve, Aviv Shamsian, Aviv Navon, Gal Chechik, and Ethan Fetaya.
\newblock Personalized federated learning with gaussian processes.
\newblock {\em Advances in Neural Information Processing Systems},
  34:8392--8406, 2021.

\bibitem[\protect\citeauthoryear{Arivazhagan \bgroup \em et al.\egroup
  }{2019}]{arivazhagan2019federated}
Manoj~Ghuhan Arivazhagan, Vinay Aggarwal, Aaditya~Kumar Singh, and Sunav
  Choudhary.
\newblock Federated learning with personalization layers.
\newblock {\em arXiv preprint arXiv:1912.00818}, 2019.

\bibitem[\protect\citeauthoryear{Chen and Chao}{2022}]{chen2021bridging}
Hong-You Chen and Wei-Lun Chao.
\newblock On bridging generic and personalized federated learning for image
  classification.
\newblock In {\em International Conference on Learning Representations}, 2022.

\bibitem[\protect\citeauthoryear{Chen \bgroup \em et al.\egroup
  }{2015}]{chen2015diffusion}
Jie Chen, C{\'e}dric Richard, and Ali~H Sayed.
\newblock Diffusion lms over multitask networks.
\newblock {\em IEEE Transactions on Signal Processing}, 63(11):2733--2748,
  2015.

\bibitem[\protect\citeauthoryear{Chen \bgroup \em et al.\egroup
  }{2022}]{ijcai2022p357}
Fengwen Chen, Guodong Long, Zonghan Wu, Tianyi Zhou, and Jing Jiang.
\newblock Personalized federated learning with a graph.
\newblock In Lud~De Raedt, editor, {\em Proceedings of the Thirty-First
  International Joint Conference on Artificial Intelligence, {IJCAI-22}}, pages
  2575--2582. International Joint Conferences on Artificial Intelligence
  Organization, 7 2022.
\newblock Main Track.

\bibitem[\protect\citeauthoryear{Chrabaszcz \bgroup \em et al.\egroup
  }{2017}]{chrabaszcz2017downsampled}
Patryk Chrabaszcz, Ilya Loshchilov, and Frank Hutter.
\newblock A downsampled variant of imagenet as an alternative to the cifar
  datasets.
\newblock {\em arXiv preprint arXiv:1707.08819}, 2017.

\bibitem[\protect\citeauthoryear{Cohen \bgroup \em et al.\egroup
  }{2017}]{cohen2017emnist}
Gregory Cohen, Saeed Afshar, Jonathan Tapson, and Andre Van~Schaik.
\newblock Emnist: Extending mnist to handwritten letters.
\newblock In {\em 2017 international joint conference on neural networks
  (IJCNN)}, pages 2921--2926. IEEE, 2017.

\bibitem[\protect\citeauthoryear{Collins \bgroup \em et al.\egroup
  }{2021}]{collins2021exploiting}
Liam Collins, Hamed Hassani, Aryan Mokhtari, and Sanjay Shakkottai.
\newblock Exploiting shared representations for personalized federated
  learning.
\newblock In {\em International Conference on Machine Learning}, pages
  2089--2099. PMLR, 2021.

\bibitem[\protect\citeauthoryear{Darlow \bgroup \em et al.\egroup
  }{2018}]{darlow2018cinic}
Luke~N Darlow, Elliot~J Crowley, Antreas Antoniou, and Amos~J Storkey.
\newblock Cinic-10 is not imagenet or cifar-10.
\newblock {\em arXiv preprint arXiv:1810.03505}, 2018.

\bibitem[\protect\citeauthoryear{Deng \bgroup \em et al.\egroup
  }{2020}]{deng2020adaptive}
Yuyang Deng, Mohammad~Mahdi Kamani, and Mehrdad Mahdavi.
\newblock Adaptive personalized federated learning.
\newblock {\em arXiv preprint arXiv:2003.13461}, 2020.

\bibitem[\protect\citeauthoryear{Fallah \bgroup \em et al.\egroup
  }{2020}]{fallah2020personalized}
Alireza Fallah, Aryan Mokhtari, and Asuman Ozdaglar.
\newblock Personalized federated learning with theoretical guarantees: A
  model-agnostic meta-learning approach.
\newblock {\em Advances in Neural Information Processing Systems},
  33:3557--3568, 2020.

\bibitem[\protect\citeauthoryear{Ghosh \bgroup \em et al.\egroup
  }{2020}]{ghosh2020efficient}
Avishek Ghosh, Jichan Chung, Dong Yin, and Kannan Ramchandran.
\newblock An efficient framework for clustered federated learning.
\newblock {\em Advances in Neural Information Processing Systems},
  33:19586--19597, 2020.

\bibitem[\protect\citeauthoryear{Hanzely and
  Richt{\'a}rik}{2020}]{hanzely2020federated}
Filip Hanzely and Peter Richt{\'a}rik.
\newblock Federated learning of a mixture of global and local models.
\newblock {\em arXiv preprint arXiv:2002.05516}, 2020.

\bibitem[\protect\citeauthoryear{He \bgroup \em et al.\egroup
  }{2020}]{he2020group}
Chaoyang He, Murali Annavaram, and Salman Avestimehr.
\newblock Group knowledge transfer: Federated learning of large cnns at the
  edge.
\newblock {\em Advances in Neural Information Processing Systems},
  33:14068--14080, 2020.

\bibitem[\protect\citeauthoryear{Hsu \bgroup \em et al.\egroup
  }{2019}]{Hsu2019MeasuringTE}
Tzu-Ming~Harry Hsu, Qi, and Matthew Brown.
\newblock Measuring the effects of non-identical data distribution for
  federated visual classification.
\newblock {\em ArXiv}, abs/1909.06335, 2019.

\bibitem[\protect\citeauthoryear{Hu \bgroup \em et al.\egroup
  }{2021}]{hu2021source}
Hongsheng Hu, Zoran Salcic, Lichao Sun, Gillian Dobbie, and Xuyun Zhang.
\newblock Source inference attacks in federated learning.
\newblock In {\em 2021 IEEE International Conference on Data Mining (ICDM)},
  pages 1102--1107. IEEE, 2021.

\bibitem[\protect\citeauthoryear{Huang \bgroup \em et al.\egroup
  }{2021}]{huang2021personalized}
Yutao Huang, Lingyang Chu, Zirui Zhou, Lanjun Wang, Jiangchuan Liu, Jian Pei,
  and Yong Zhang.
\newblock Personalized cross-silo federated learning on non-iid data.
\newblock In {\em AAAI}, pages 7865--7873, 2021.

\bibitem[\protect\citeauthoryear{Jin \bgroup \em et al.\egroup
  }{2020}]{jin2020affine}
Danqi Jin, Jie Chen, C{\'e}dric Richard, Jingdong Chen, and Ali~H Sayed.
\newblock Affine combination of diffusion strategies over networks.
\newblock {\em IEEE Transactions on Signal Processing}, 68:2087--2104, 2020.

\bibitem[\protect\citeauthoryear{Karimireddy \bgroup \em et al.\egroup
  }{2020}]{karimireddy2020scaffold}
Sai~Praneeth Karimireddy, Satyen Kale, Mehryar Mohri, Sashank Reddi, Sebastian
  Stich, and Ananda~Theertha Suresh.
\newblock Scaffold: Stochastic controlled averaging for federated learning.
\newblock In {\em International Conference on Machine Learning}, pages
  5132--5143. PMLR, 2020.

\bibitem[\protect\citeauthoryear{Khodak \bgroup \em et al.\egroup
  }{2019}]{DBLP:conf/nips/KhodakBT19}
Mikhail Khodak, Maria-Florina Balcan, and Ameet~S. Talwalkar.
\newblock Adaptive gradient-based meta-learning methods.
\newblock In {\em NeurIPS}, 2019.

\bibitem[\protect\citeauthoryear{Krizhevsky \bgroup \em et al.\egroup
  }{2009}]{krizhevsky2009learning}
Alex Krizhevsky, Geoffrey Hinton, et~al.
\newblock Learning multiple layers of features from tiny images.
\newblock {\em Tech Report}, 2009.

\bibitem[\protect\citeauthoryear{Li \bgroup \em et al.\egroup
  }{2019}]{li2019convergence}
Xiang Li, Kaixuan Huang, Wenhao Yang, Shusen Wang, and Zhihua Zhang.
\newblock On the convergence of fedavg on non-iid data.
\newblock {\em arXiv preprint arXiv:1907.02189}, 2019.

\bibitem[\protect\citeauthoryear{Li \bgroup \em et al.\egroup
  }{2020a}]{li2020federated}
Tian Li, Anit~Kumar Sahu, Manzil Zaheer, Maziar Sanjabi, Ameet Talwalkar, and
  Virginia Smith.
\newblock Federated optimization in heterogeneous networks.
\newblock {\em Proceedings of Machine Learning and Systems}, 2:429--450, 2020.

\bibitem[\protect\citeauthoryear{Li \bgroup \em et al.\egroup
  }{2020b}]{Li2020OnTC}
Xiang Li, Kaixuan Huang, Wenhao Yang, Shusen Wang, and Zhihua Zhang.
\newblock On the convergence of fedavg on non-iid data.
\newblock {\em ArXiv}, abs/1907.02189, 2020.

\bibitem[\protect\citeauthoryear{Li \bgroup \em et al.\egroup
  }{2021a}]{li2021model}
Qinbin Li, Bingsheng He, and Dawn Song.
\newblock Model-contrastive federated learning.
\newblock In {\em Proceedings of the IEEE/CVF Conference on Computer Vision and
  Pattern Recognition}, pages 10713--10722, 2021.

\bibitem[\protect\citeauthoryear{Li \bgroup \em et al.\egroup
  }{2021b}]{li2021ditto}
Tian Li, Shengyuan Hu, Ahmad Beirami, and Virginia Smith.
\newblock Ditto: Fair and robust federated learning through personalization.
\newblock In {\em International Conference on Machine Learning}, pages
  6357--6368. PMLR, 2021.

\bibitem[\protect\citeauthoryear{Li \bgroup \em et al.\egroup
  }{2022}]{li2022learning}
Shuangtong Li, Tianyi Zhou, Xinmei Tian, and Dacheng Tao.
\newblock Learning to collaborate in decentralized learning of personalized
  models.
\newblock In {\em Proceedings of the IEEE/CVF Conference on Computer Vision and
  Pattern Recognition}, pages 9766--9775, 2022.

\bibitem[\protect\citeauthoryear{Mansour \bgroup \em et al.\egroup
  }{2020}]{Mansour2020ThreeAF}
Y.~Mansour, Mehryar Mohri, Jae Ro, and Ananda~Theertha Suresh.
\newblock Three approaches for personalization with applications to federated
  learning.
\newblock {\em ArXiv}, abs/2002.10619, 2020.

\bibitem[\protect\citeauthoryear{Marfoq \bgroup \em et al.\egroup
  }{2021}]{marfoq2021federated}
Othmane Marfoq, Giovanni Neglia, Aur{\'e}lien Bellet, Laetitia Kameni, and
  Richard Vidal.
\newblock Federated multi-task learning under a mixture of distributions.
\newblock {\em Advances in Neural Information Processing Systems},
  34:15434--15447, 2021.

\bibitem[\protect\citeauthoryear{Marfoq \bgroup \em et al.\egroup
  }{2022}]{marfoq2022personalized}
Othmane Marfoq, Giovanni Neglia, Richard Vidal, and Laetitia Kameni.
\newblock Personalized federated learning through local memorization.
\newblock In {\em International Conference on Machine Learning}, pages
  15070--15092. PMLR, 2022.

\bibitem[\protect\citeauthoryear{McMahan \bgroup \em et al.\egroup
  }{2017}]{mcmahan2017communication}
Brendan McMahan, Eider Moore, Daniel Ramage, Seth Hampson, and Blaise~Aguera
  y~Arcas.
\newblock Communication-efficient learning of deep networks from decentralized
  data.
\newblock In {\em Artificial intelligence and statistics}, pages 1273--1282.
  PMLR, 2017.

\bibitem[\protect\citeauthoryear{Mills \bgroup \em et al.\egroup
  }{2022}]{Mills2022MultiTaskFL}
Jed Mills, Jia Hu, and Geyong Min.
\newblock Multi-task federated learning for personalised deep neural networks
  in edge computing.
\newblock {\em IEEE Transactions on Parallel and Distributed Systems},
  33:630--641, 2022.

\bibitem[\protect\citeauthoryear{Oh \bgroup \em et al.\egroup
  }{2021}]{oh2021fedbabu}
Jaehoon Oh, Sangmook Kim, and Se-Young Yun.
\newblock Fedbabu: Towards enhanced representation for federated image
  classification.
\newblock {\em arXiv preprint arXiv:2106.06042}, 2021.

\bibitem[\protect\citeauthoryear{Sahu \bgroup \em et al.\egroup
  }{2020}]{Sahu2020FederatedOI}
Anit~Kumar Sahu, Tian Li, Maziar Sanjabi, Manzil Zaheer, Ameet~S. Talwalkar,
  and Virginia Smith.
\newblock Federated optimization in heterogeneous networks.
\newblock {\em arXiv: Learning}, 2020.

\bibitem[\protect\citeauthoryear{Sattler \bgroup \em et al.\egroup
  }{2021}]{Sattler2021ClusteredFL}
Felix Sattler, Klaus-Robert M{\"u}ller, and Wojciech Samek.
\newblock Clustered federated learning: Model-agnostic distributed multitask
  optimization under privacy constraints.
\newblock {\em IEEE Transactions on Neural Networks and Learning Systems},
  32:3710--3722, 2021.

\bibitem[\protect\citeauthoryear{Shamsian \bgroup \em et al.\egroup
  }{2021}]{shamsian2021personalized}
Aviv Shamsian, Aviv Navon, Ethan Fetaya, and Gal Chechik.
\newblock Personalized federated learning using hypernetworks.
\newblock In {\em International Conference on Machine Learning}, pages
  9489--9502. PMLR, 2021.

\bibitem[\protect\citeauthoryear{Shin \bgroup \em et al.\egroup
  }{2020}]{shin2020xor}
MyungJae Shin, Chihoon Hwang, Joongheon Kim, Jihong Park, Mehdi Bennis, and
  Seong-Lyun Kim.
\newblock Xor mixup: Privacy-preserving data augmentation for one-shot
  federated learning.
\newblock {\em arXiv preprint arXiv:2006.05148}, 2020.

\bibitem[\protect\citeauthoryear{Smith \bgroup \em et al.\egroup
  }{2017}]{DBLP:conf/nips/SmithCST17}
Virginia Smith, Chao-Kai Chiang, Maziar Sanjabi, and Ameet~S. Talwalkar.
\newblock Federated multi-task learning.
\newblock In {\em NIPS}, 2017.

\bibitem[\protect\citeauthoryear{T~Dinh \bgroup \em et al.\egroup
  }{2020}]{t2020personalized}
Canh T~Dinh, Nguyen Tran, and Josh Nguyen.
\newblock Personalized federated learning with moreau envelopes.
\newblock {\em Advances in Neural Information Processing Systems},
  33:21394--21405, 2020.

\bibitem[\protect\citeauthoryear{Tan \bgroup \em et al.\egroup
  }{2022}]{tan2022fedproto}
Yue Tan, Guodong Long, Lu~Liu, Tianyi Zhou, Qinghua Lu, Jing Jiang, and Chengqi
  Zhang.
\newblock Fedproto: Federated prototype learning across heterogeneous clients.
\newblock In {\em AAAI Conference on Artificial Intelligence}, volume~1,
  page~3, 2022.

\bibitem[\protect\citeauthoryear{Wang \bgroup \em et al.\egroup
  }{2019}]{wang2019federated}
Kangkang Wang, Rajiv Mathews, Chlo{\'e} Kiddon, Hubert Eichner, Fran{\c{c}}oise
  Beaufays, and Daniel Ramage.
\newblock Federated evaluation of on-device personalization.
\newblock {\em arXiv preprint arXiv:1910.10252}, 2019.

\bibitem[\protect\citeauthoryear{Yoon \bgroup \em et al.\egroup
  }{2021}]{yoon2021fedmix}
Tehrim Yoon, Sumin Shin, Sung~Ju Hwang, and Eunho Yang.
\newblock Fedmix: Approximation of mixup under mean augmented federated
  learning.
\newblock {\em arXiv preprint arXiv:2107.00233}, 2021.

\bibitem[\protect\citeauthoryear{Yu \bgroup \em et al.\egroup
  }{2020}]{Yu2020SalvagingFL}
Tao Yu, Eugene Bagdasaryan, and Vitaly Shmatikov.
\newblock Salvaging federated learning by local adaptation.
\newblock {\em ArXiv}, abs/2002.04758, 2020.

\bibitem[\protect\citeauthoryear{Yue \bgroup \em et al.\egroup
  }{2021}]{DBLP:conf/mobihoc/YueRXLZ21}
Sheng Yue, Ju~Ren, Jiang Xin, Sen Lin, and Junshan Zhang.
\newblock Inexact-admm based federated meta-learning for fast and continual
  edge learning.
\newblock {\em Proceedings of the Twenty-second International Symposium on
  Theory, Algorithmic Foundations, and Protocol Design for Mobile Networks and
  Mobile Computing}, 2021.

\bibitem[\protect\citeauthoryear{Zhang \bgroup \em et al.\egroup
  }{2021a}]{zhang2021parameterized}
Jie Zhang, Song Guo, Xiaosong Ma, Haozhao Wang, Wenchao Xu, and Feijie Wu.
\newblock Parameterized knowledge transfer for personalized federated learning.
\newblock {\em Advances in Neural Information Processing Systems},
  34:10092--10104, 2021.

\bibitem[\protect\citeauthoryear{Zhang \bgroup \em et al.\egroup
  }{2021b}]{Zhang2021PersonalizedFL}
Michael Zhang, Karan Sapra, Sanja Fidler, Serena Yeung, and Jos{\'e}~Manuel
  {\'A}lvarez.
\newblock Personalized federated learning with first order model optimization.
\newblock {\em ArXiv}, abs/2012.08565, 2021.

\bibitem[\protect\citeauthoryear{Zhao and Sayed}{2012}]{zhao2012clustering}
Xiaochuan Zhao and Ali~H Sayed.
\newblock Clustering via diffusion adaptation over networks.
\newblock In {\em 2012 3rd International Workshop on Cognitive Information
  Processing (CIP)}, pages 1--6. IEEE, 2012.

\bibitem[\protect\citeauthoryear{Zhao \bgroup \em et al.\egroup
  }{2018}]{zhao2018federated}
Yue Zhao, Meng Li, Liangzhen Lai, Naveen Suda, Damon Civin, and Vikas Chandra.
\newblock Federated learning with non-iid data.
\newblock {\em arXiv preprint arXiv:1806.00582}, 2018.

\end{thebibliography}

\newpage
\clearpage
\appendix

\section{Theoretical Results}\label{Theoretical_analysis}

% \subsection{Assumptions of risk functions}
% \label{appendix:assumptions}
% \begin{definition}
%     ($L$-Lipschitz continuous gradient). A differentiable convex function $f$ is said to have an $L$-Lipschitz continuous gradient, if there exists a constant $L>0$, such that:
%     \begin{equation}
%         \left\| \nabla f(x) - \nabla f(y) \right\| \leq L \left\| x - y \right\|, \forall x, y
%     \end{equation}
%     If $f$ has an $L$-Lipschitz continuous gradient, then it holds that:
%     \begin{equation}
%         f(y) \leq f(x) + \left \langle \nabla f(x), y-x  \right \rangle + \frac{L}{2} \left\| y-x\right\|^2, \forall x, y
%     \end{equation}
% \end{definition}

% \begin{definition}
%     ($c$-strongly convex). A differentiable convex function $f$ is said to be $c$-strongly convex if there exists a constant $c>0$, such that:
%     \begin{equation}
%         f(y) \ge f(x) + \left \langle \nabla f(x), y-x  \right \rangle + \frac{c}{2} \left\| y-x\right\|^2, \forall x, y
%     \end{equation}
% \end{definition}

% If a function $f$ is $c$-strongly convex and has an $L$-Lipschitz continuous graident, then it is obvious  that $c \leq L$.

\subsection{A Tony Example for Problem \eqref{EQ:ConvertedOpt}}
\label{appendix:tony_example}
If the matrix $\mathbf{W}_{i}$ is not invertible and has a rank less than $N-1$, the solution of Eq.~\eqref{EQ:ConvertedOpt} will be not unique. For example, consider the case when $N=3$, and 
\begin{equation*}
    \mathbf{W}_{i} = \begin{pmatrix}
 0 & 0 & 0\\
 0 & 0 & 0\\
 0 & 0 & 1
\end{pmatrix}
\end{equation*}
It means that the model of client $i$ is similar to that of clients 1 and 2, and is very different from that of client 3. In this case,  any vector $\mathbf{p}_{i}=(p_{i,1}, p_{i,2}, 0)^T$ with $p_{i,1}+p_{i,2}=1$ is the solution of Eq.~\eqref{EQ:ConvertedOpt}. 
For example, $\{p_{i,1}=0.9, p_{i,2}=0.1\}$ and  $\{p_{i,1}=0.1, p_{i,2}=0.9\}$ are two sets of solutions that satisfy the requirements respectively. But, it is obvious that we will infer different things from these two different sets of solutions. In the first set of solutions, we will think that client $i$ is similar to client 1, not to similar to client 2; In the second set of solutions, we will think that client $i$ is similar to client 2, not so similar to client 1. Therefore, directly solving Eq.~\eqref{EQ:ConvertedOpt} in this case  does not give us a unique solution.

\subsection{Optimal Solution of Problem \eqref{solution2}}
\label{appendix:solution_p}
To solve the following optimization problem:
\begin{equation}
\begin{aligned}
    &\min_{\mathbf{p}_{i}}\quad \sum_{j=1}^{N}p_{i,j}^2\left\| \hat{w_{i}}^{\star}-\hat{w}_{j}^{t}\right\|^{2}, \\
    &\text{subject to} \quad \mathbf{1}^{T}_{N} \mathbf{p}_{i}=1, p_{i,j} \ge 0.
\end{aligned}
\label{appendix_1}
\end{equation}
First we construct the Lagrangian function with respect to the equality constraint and discard the non-negativity constraint.
\begin{equation}
\begin{aligned}
    L(\lambda) = &\sum_{j=1}^{N}p_{i,j}^2\left\| \hat{w_{i}}^{\star}-\hat{w}_{j}^{t}\right\|^{2} +\lambda( \mathbf{1}^{T}_{N} \mathbf{p}_{i}-1)
\end{aligned}
\end{equation}
Then, take the derivative of $\lambda$ and $p_{i,j}$. respectively, and we get:
\begin{equation}
\begin{aligned}
    2p_{i,j}\left\| \hat{w_{i}}^{\star}-\hat{w}_{j}^{t}\right\|^{2} - \lambda = 0
\end{aligned}
\end{equation}
\begin{equation}
\begin{aligned}
    \mathbf{1}^{T}_{N} \mathbf{p}_{i} -1=0
\end{aligned}
\end{equation}
It turns out that the value of $\lambda$ is:
\begin{equation}
\begin{aligned}
    \lambda = \frac{2}{\sum_{j=1}^{N}\left\| \hat{w_{i}}^{\star}-\hat{w}_{j}^{t}\right\|^{-2}}
\end{aligned}
\end{equation}
Finally, we can determine that:
\begin{equation}
    p_{i,j}=\frac{\left\| \hat{w_{i}}^{\star}-\hat{w}_{j}^{t}\right\|^{-2}}{\sum_{k=1}^{N}\left\| \hat{w_{i}}^{\star}-\hat{w}_{k}^{t}\right\|^{-2}}.
    \label{appendix_solution}
\end{equation}
Given that it Eq.~\eqref{appendix_solution} satisfies  the non-negativity constraint $p_{i,j}\ge 0$, therefore, Eq.~\eqref{appendix_solution} is the solution of Eq.~\eqref{appendix_1}.

\section{Details of Experiment Setup}
We implement our method and other baseline methods in PyTorch 1.7, and the simulation server is equipped with a Tesla V100 GPU, a 2.4-GHz Inter Core E5-2680 CPU and 256GB of memory. 
\label{appendix:experiment_setup}

\subsection{Datasets and Models}
\label{appendix: datasets_models}
We consider image classification tasks and evaluate our method on five popular datasets: (1) EMNIST (Extend MNIST) is a 62-class image classification dataset, extending the classic MNIST dataset. It contains 62 categories of handwritten characters, including 10 digits, 26 uppercase letters and 26 lowercase letters. There are 814,255 images in total; (2) CIFAR10 dataset consists of 60,000 32x32 colour images in 10 classes, with 6,000 images per class. There are 50,000 training images and 10,000 test images; (3) CIFAR100 dataset consists of 60,000 32x32 colour images in 100 classes, with 600 images per class. There are 500 training images and 100 test images; (4) CINIC-10, which is more diverse than CIFAR10 as it is constructed from two different sources: ImageNet and CIFAR10. This dataset consists of 100,000 32x32 colour images in 10 classes; (5) Tiny-ImageNet is constructed from ImageNet and it consists of 100,000 32x32 colour images in 200 classes. We construct three different CNN models for classifying EMNIST, CIFAR10/CIFAR100/CINIC-10 and Tiny-ImageNet images, respectively. The first CNN model is constructed by two convolutional layers followed by pooling, and two fully connected layers with a final dense layer containing 2,048 units. The second CNN model has two convolutional pooling layers, two batch normalization layers and two fully connected layers, and the rate of Dropout is set to 0.5. The third CNN model is ResNet-8, and the architecture is the same as \cite{he2020group}.
\subsection{Data Partitioning}
 We simulate the heterogeneous settings with three widely used scenarios, including a pathological setting and two practical settings. 
 \begin{itemize}
    \item {\bf  Pathological Heterogeneous Setting.}  Each client is randomly assigned with a small number of classes of samples \cite{mcmahan2017communication,shamsian2021personalized}. We sample 4, 2, 2, 6 and 10 classes for EMNIST, CIFAR10, CINIC10, CIFAR100, Tiny-ImageNet from a total of 62, 10, 10, 100, 200 classes for each client, respectively. There is no group-wise similarity between clients in this setting.
    \item  {\bf Practical Heterogeneous Setting 1.}  All clients have the same data size but different distributions. For each client, $s\%$ of data ($80\%$ by default) are selected from a set of dominant classes, and the remaining $(100-s)\%$ are uniformly sampled from all classes \cite{karimireddy2020scaffold,huang2021personalized}. All clients are divided into multiple groups. Clients in each group share the same dominant classes implying that there is an underlying clustering structure between clients. Specifically, for CIFAR10 and CINIC-10 datasets which have 10 categories of images, we divide the clients into 4 groups and the number of dominant class for each client in the same group is 3. For CIFAR100 dataset which has 100 categories of images, we divide the clients into 4 groups and the number of dominant class for each client in the same group is 20; For Tiny-ImageNet dataset which have 200 categories, we divide the clients into 4 groups and the number of dominant class for each client in the same group is 40.
    \item {\bf Practical Heterogeneous Setting 2.}   Each client contains most of the classes but the data in each class is not uniformly distributed \cite{Hsu2019MeasuringTE,li2021model,chen2021bridging}. We create the federated version by randomly partitioning datasets among $N$ clients using a symmetric Dirichlet distribution $\text{Dir}(\alpha)$ ($\alpha=0.07$ by default). For example, for each class $c$,  we sample a vector $p_{c}$ from $\text{Dir}(\alpha)$ and allocate to client $m$ a  fraction $p_{c,m}$ of all training instances of class $c$.
\end{itemize}

For each setting, the test data on each client has the same distribution as that of the training data.
% {\bf YP: no idea how you can set 4:1 since the fraction of test samples in each dataset is a fixed number as you have described. }
\subsection{Implementation Details of Methods}
\label{appendix: implementation_of_baselines}
By default, we set  $K$  as 5. When tuning $K$, we may change it from $1-10$. For FedProx\footnote{https://github.com/litian96/FedProx}, we search $\mu$ from   $\{0.01,0.1,1,10\}$ to find its best value  1. For FedAvgM, we search the momentum value $\beta$ of Nesterov accelerated gradient from  $\{0,0.1,0.5,0.9,0.99\}$ and find its best value  0.1. For pFedMe\footnote{https://github.com/CharlieDinh/pFedMe}, we search $\lambda$ from  $\{0.1,1,10,100\}$ to find its best value 1.  The local iterations $K$ to find a $\delta$-approximation personalized models is set to 5, and the coefficient 
 of smooth aggregation $\beta$ is set to 1, which is the same as the  original paper. For FedFomo\footnote{https://github.com/NVlabs/FedFomo}, we set the number of models downloaded as $M=5$ which is recommended in the paper; For ClusterFL\footnote{https://github.com/felisat/clustered-federated-learning}, we use the same values of tolerance as the ones used in its official implementation for EMNIST dataset, and  the hyper-parameters of this algorithm are fine-tuned on the other datasets.
 % {\bf YP: then what? how you tune tol1 and tol2?}  
 For FedAMP, we employ  the grid search technology to tune the hyper-parameters, and finally the hyper-parameters are set as: $\sigma=1, \alpha=1$ and $\lambda=0.1$. Here, we use the same symbols as those in the original paper. For FedRoD, we use the last linear layer as the Personalized-head layer. For SFL\footnote{https://github.com/dawenzi098/SFL-Structural-Federated-Learning}, we set the same hyper-parameter settings as these in its official implementation. For the implementation of FedAMP and FedRoD, we make use of open source libraries \footnote{https://github.com/TsingZ0/PFL-Non-IID}. 
\section{Additional Experimental Results}
\label{appendix:additional_experiments}
\subsection{Personalized Weighting}
\label{appendix:additional_personalized_weights}
\begin{figure}[htpb]
\centering
\subfigure[Support for different numbers of distributions using EMNIST dataset.]{
    % \label{fig:iter_l} 
    \includegraphics[width=0.8\columnwidth]{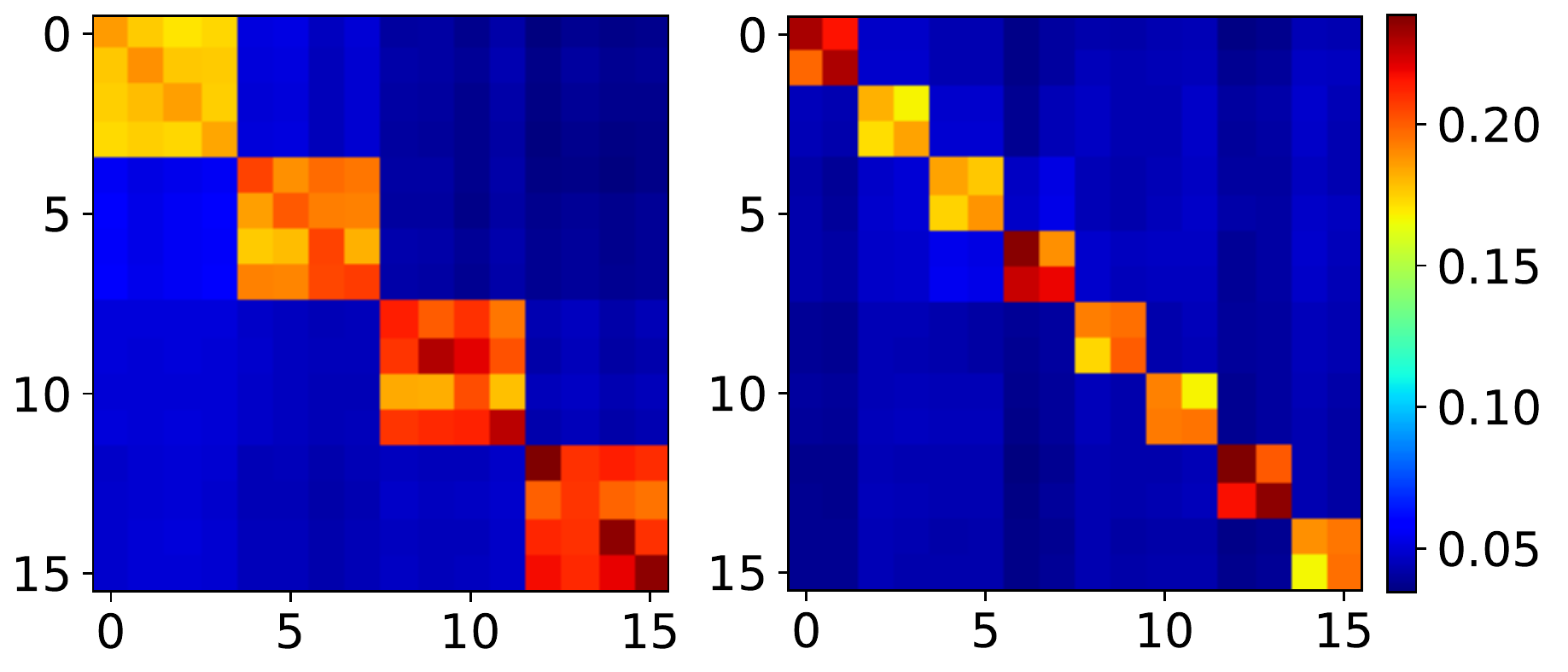}% Reduce the figure size so that it is slightly narrower than the column. Don't use precise values for figure width.This setup will avoid overfull boxes.
\label{appendix: emnist_different_groups}
}

\subfigure[Robustness to number of clients using EMNIST dataset.]{
    \includegraphics[width=0.8\columnwidth]{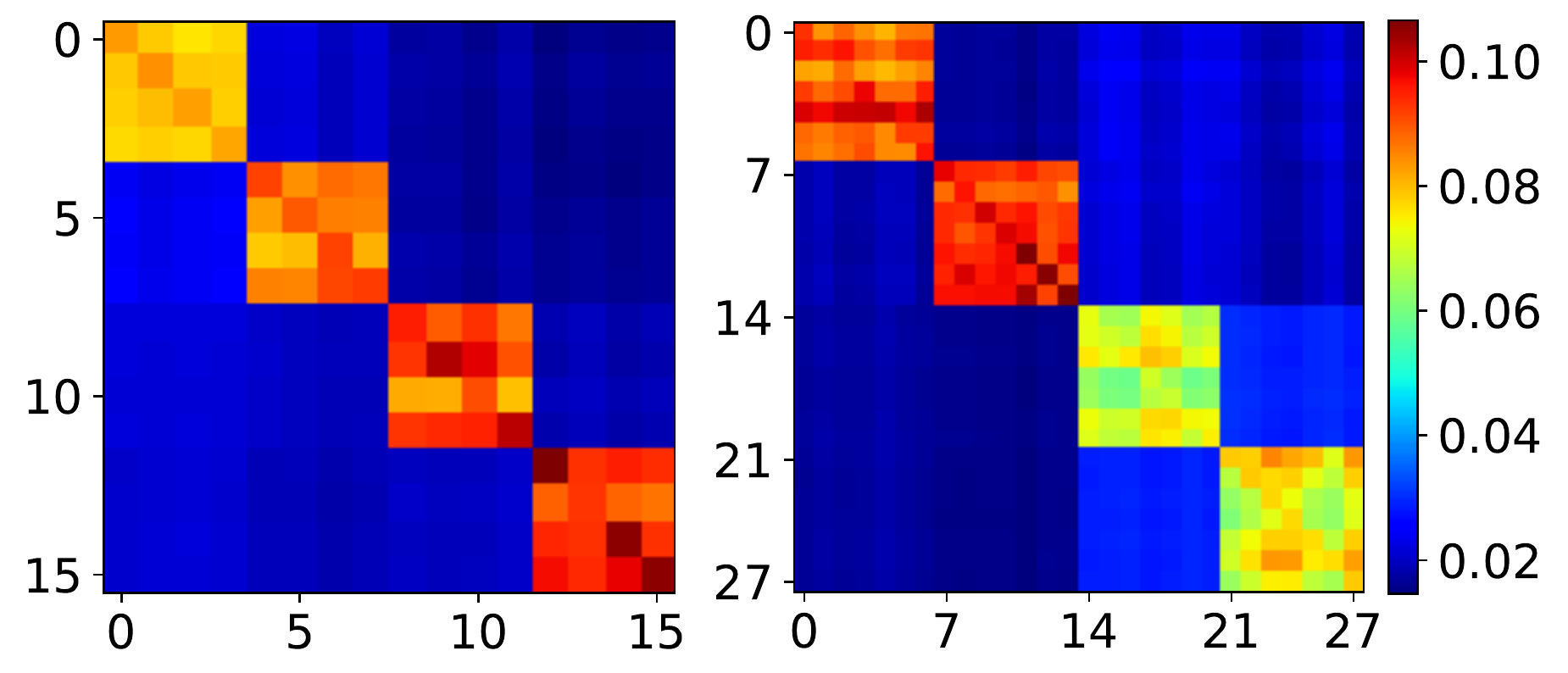}% Reduce the figure size so that it is slightly narrower than the column. Don't use precise values for figure width.This setup will avoid overfull boxes.
\label{appendix: emnist_different_numbers}
}

\subfigure[Support for different numbers of distributions using CIFAR10 dataset.]{
    \includegraphics[width=0.8\columnwidth]{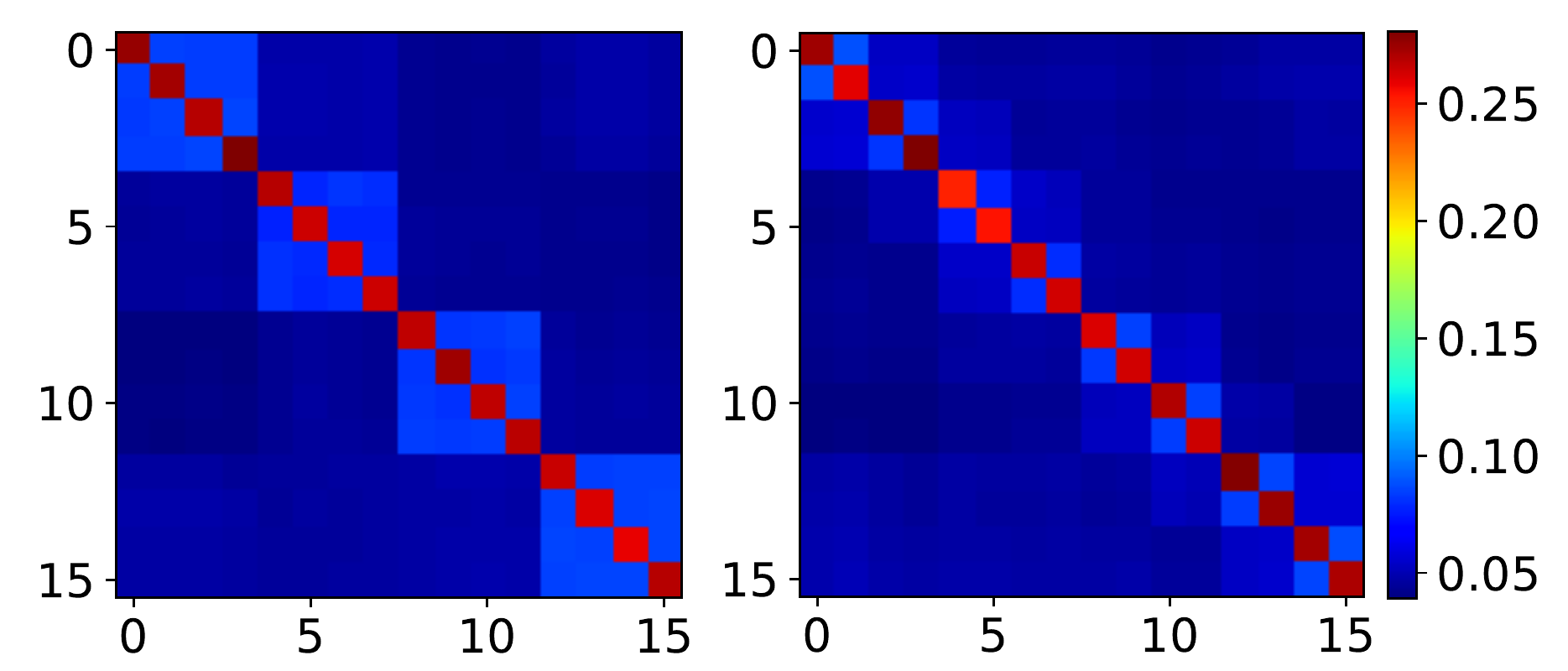}% Reduce the figure size so that it is slightly narrower than the column. Don't use precise values for figure width.This setup will avoid overfull boxes.
\label{appendix: cifar10_different_groups}
}

\subfigure[Robustness to number of clients using CIFAR10 dataset.]{
    \includegraphics[width=0.8\columnwidth]{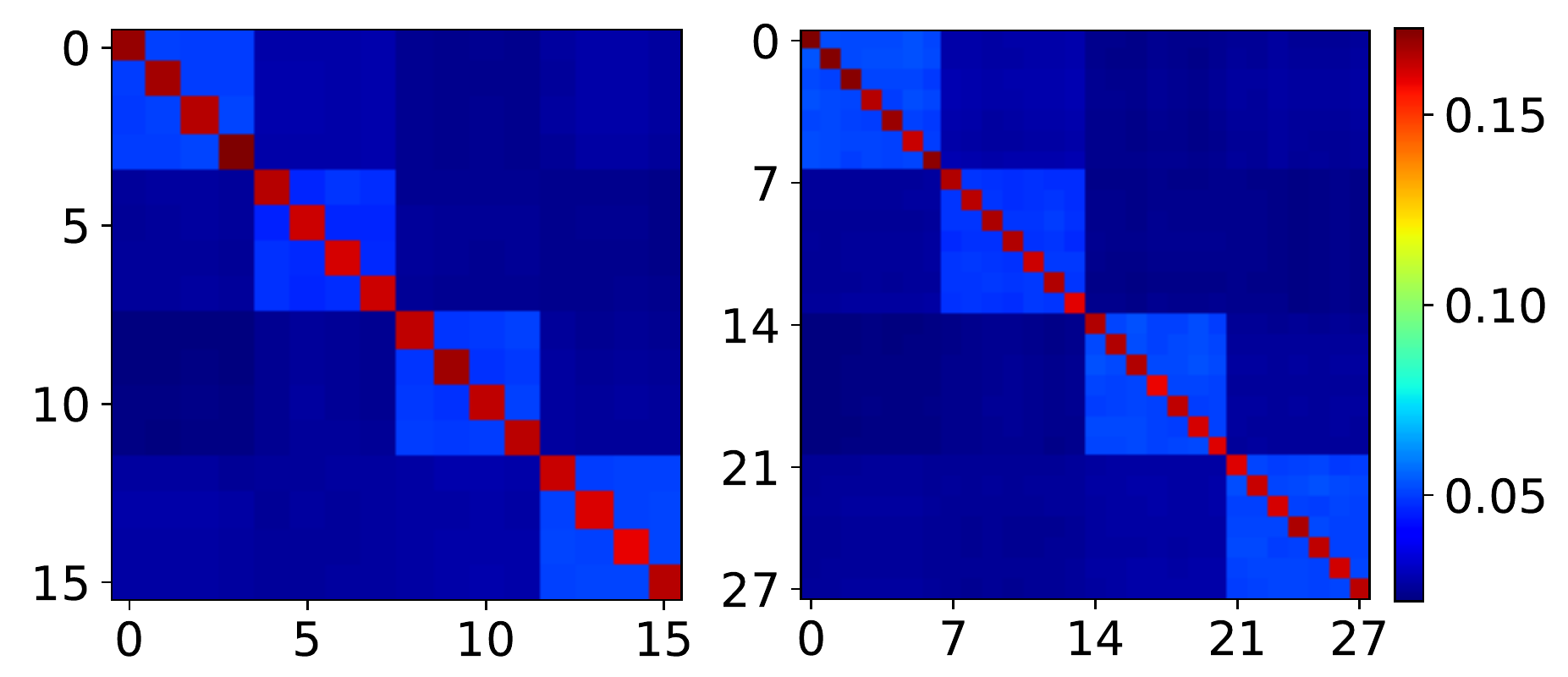}% Reduce the figure size so that it is slightly narrower than the column. Don't use precise values for figure width.This setup will avoid overfull boxes.
\label{appendix: cifar10_different_numbers}
}
\caption{The visualization of the aggregation weights $p_{i,k}$ computed by FedDWA on EMNIST and CIFAR10 datasets. The x-axis and y-axis means the IDs of clients.}
\label{appendix: personalized_weights}
\end{figure}

In order to show how FedDWA allows clients to find their optimal personalized models by properly selecting other  clients, we next visualize the aggregation weights $p_{i,k}$ computed by FedDWA. We utilize the practical heterogeneous setting 1 in which  clients are divided into multiple groups. The data distribution of clients in the same group is similar, while the data distribution of clients in different groups is different. Specifically, for clients in the same group,  $80\%$ of data samples of every client are uniformly sampled from a set of dominating classes, and $20\%$ of data samples are uniformly sampled from the rest of classes. 
% Therefore, each client has data from all classes, but the number of samples for each class is different. 
We depict clients with the same local data distributions next to each other (e.g. clients 0-4 are in the same group that have similar data distribution and client 5-9 are in the same next group). In Figure \ref{appendix: personalized_weights}, we show the aggregation weights $p_{i,k}$ computed by FedDWA using EMNIST and CIFAR10 datasets. We test the stability of FedDWA with different clients and different data distributions. The experimental results show that our method is still effective in the case of different number of clients and different number of clusters.
% {\bf YP: what is robustness? no clear definition.  what is purpose of this experiment? just showing robustness? }

\begin{figure*}[t]
\centering
\subfigure[CIFAR10]{
    % \label{fig:iter_l} 
    \includegraphics[width=0.65\columnwidth]{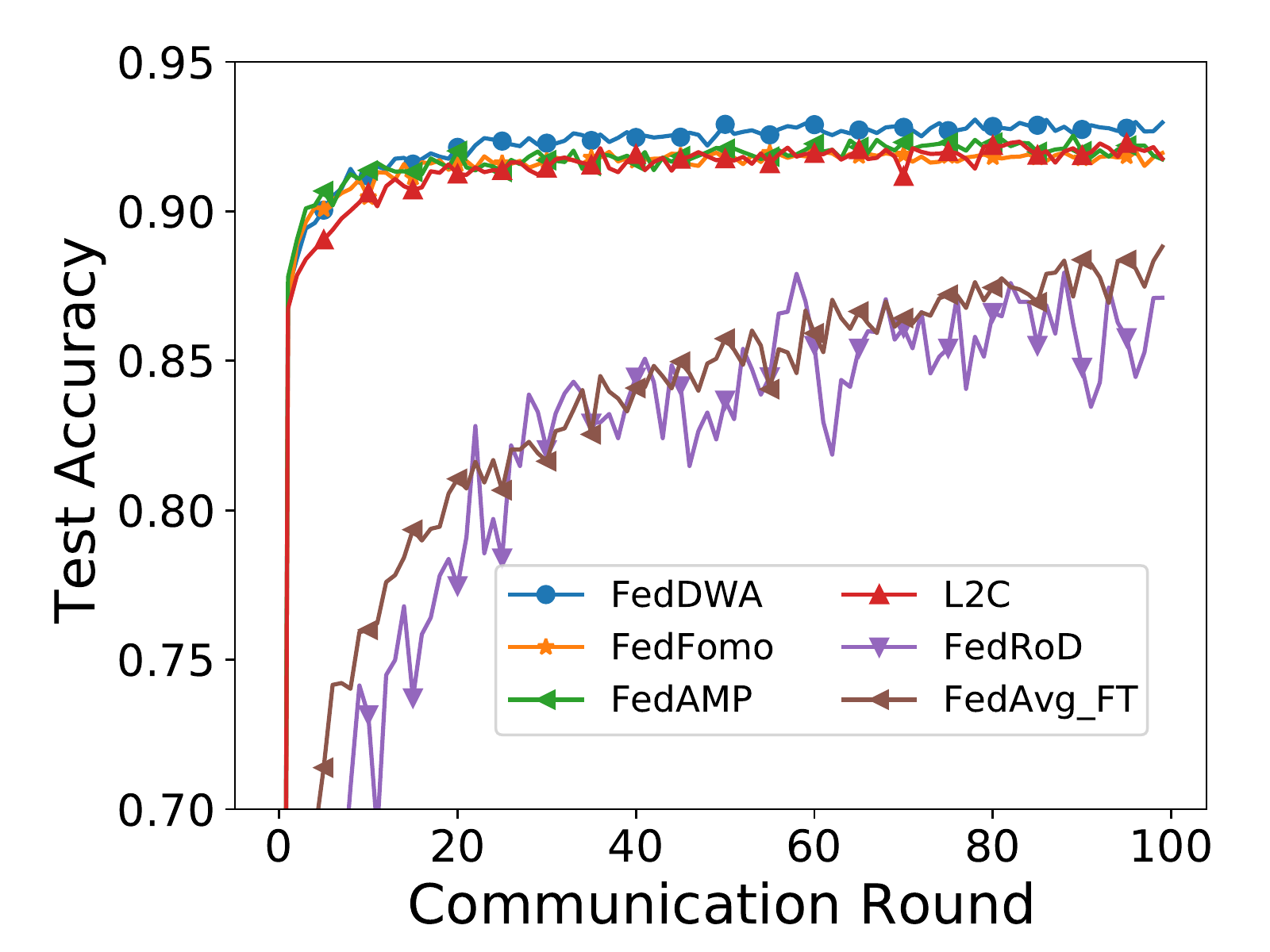}% Reduce the figure size so that it is slightly narrower than the column. Don't use precise values for figure width.This setup will avoid overfull boxes.
\label{appendix: appendix_cifar10_noniid8_curve}
}
\subfigure[CIFAR100]{
    \includegraphics[width=0.65\columnwidth]{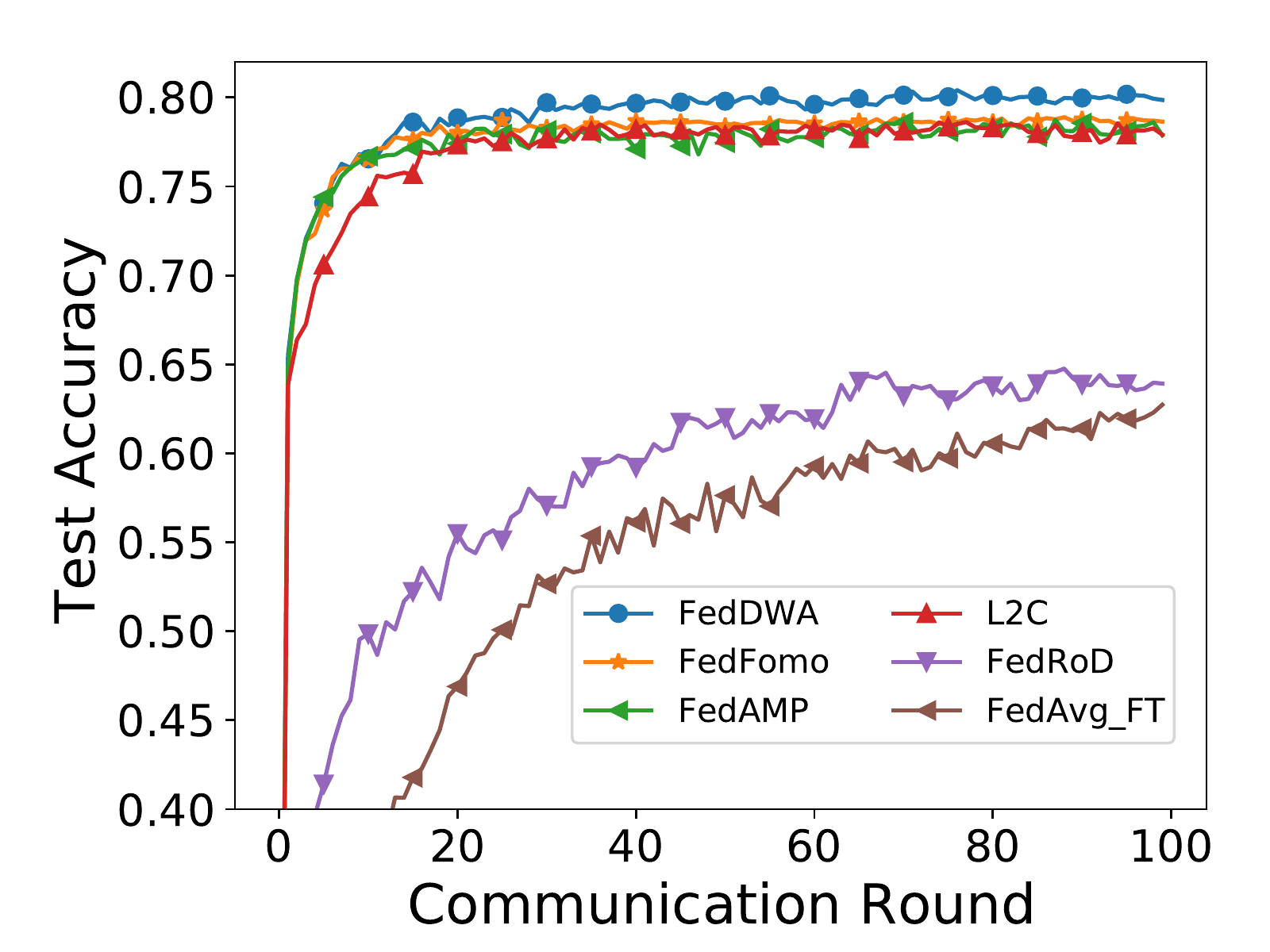}% Reduce the figure size so that it is slightly narrower than the column. Don't use precise values for figure width.This setup will avoid overfull boxes.
\label{appendix: appendix_cifar100_noniid8_curve}
}
\subfigure[CINIC10]{
    \includegraphics[width=0.65\columnwidth]{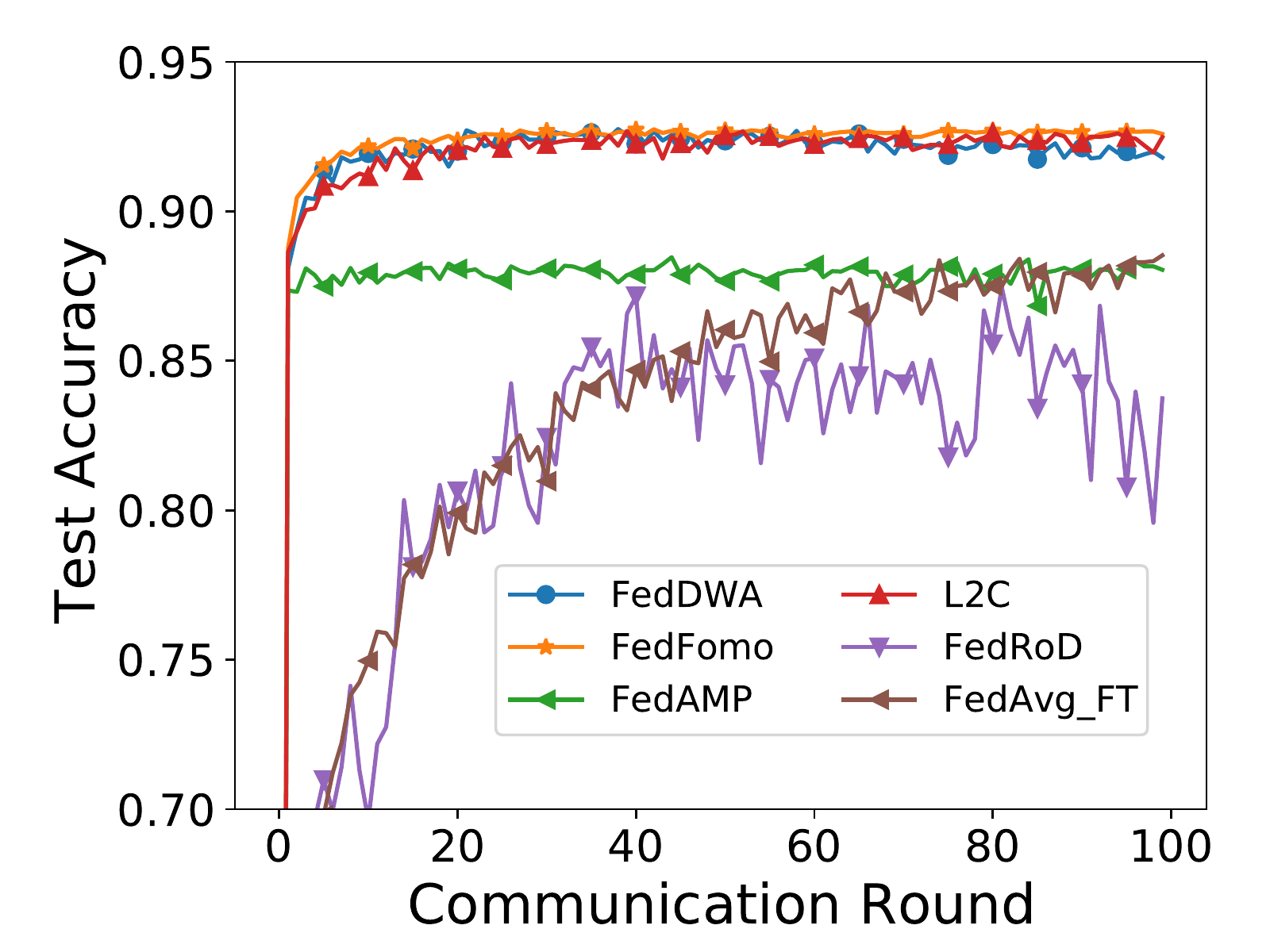}% Reduce the figure size so that it is slightly narrower than the column. Don't use precise values for figure width.This setup will avoid overfull boxes.
\label{appendix: appendix_cinic10_noniid8_curve}
}
\caption{Test accuracy over communication rounds under the pathological heterogeneous setting with 20 clients.}
\label{appendix: appendix_noniid8_curve}
\end{figure*}

\begin{figure*}[t]
\centering
\subfigure[CIFAR10]{
    \includegraphics[width=0.65\columnwidth]{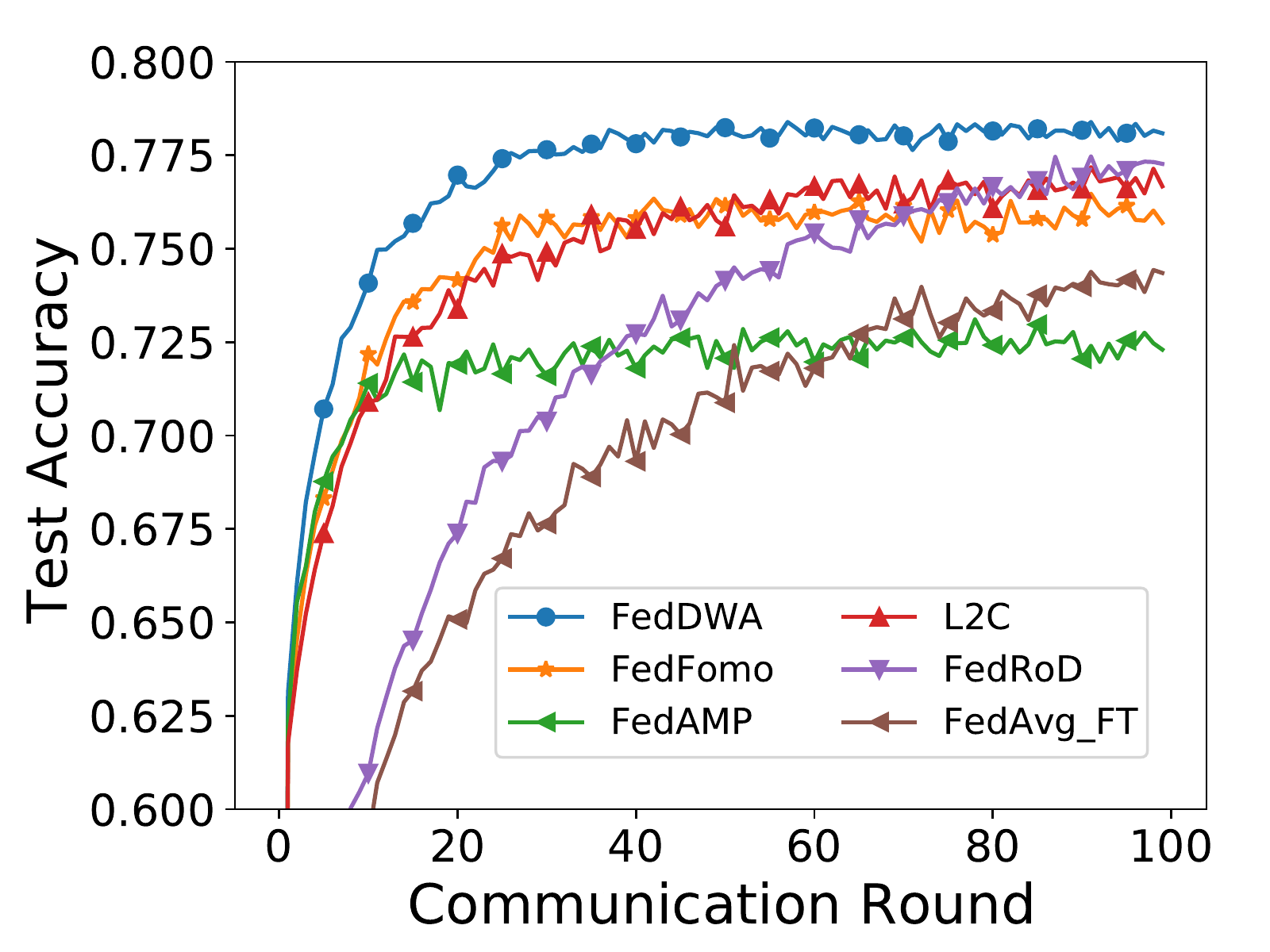}% Reduce the figure size so that it is slightly narrower than the column. Don't use precise values for figure width.This setup will avoid overfull boxes.
\label{appendix: appendix_cifar10_noniid10_curve}
}
\subfigure[CIFAR100]{
    \includegraphics[width=0.65\columnwidth]{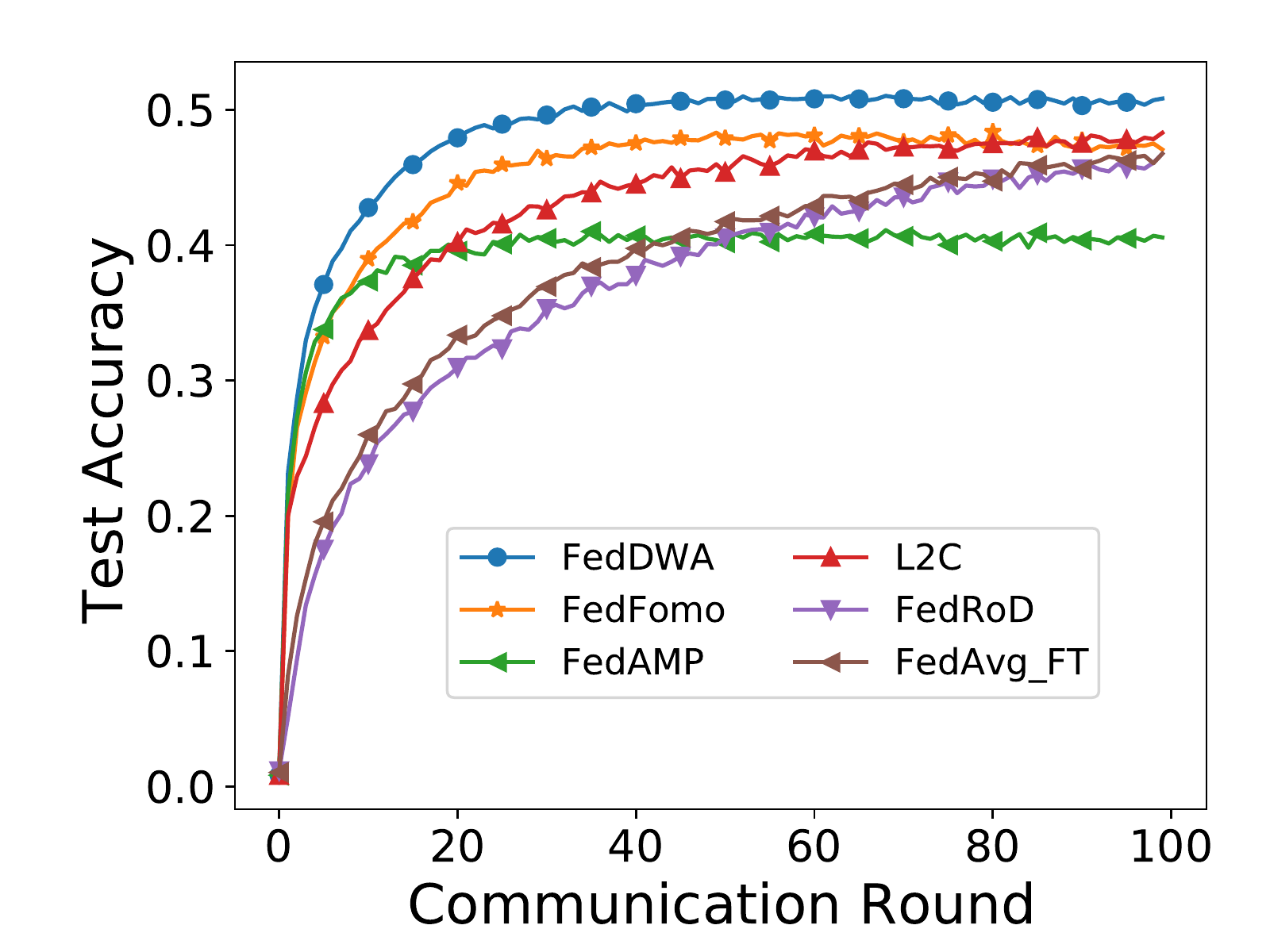}% Reduce the figure size so that it is slightly narrower than the column. Don't use precise values for figure width.This setup will avoid overfull boxes.
\label{appendix: appendix_cifar100_noniid10_curve}
}
\subfigure[CINIC10]{
    \includegraphics[width=0.65\columnwidth]{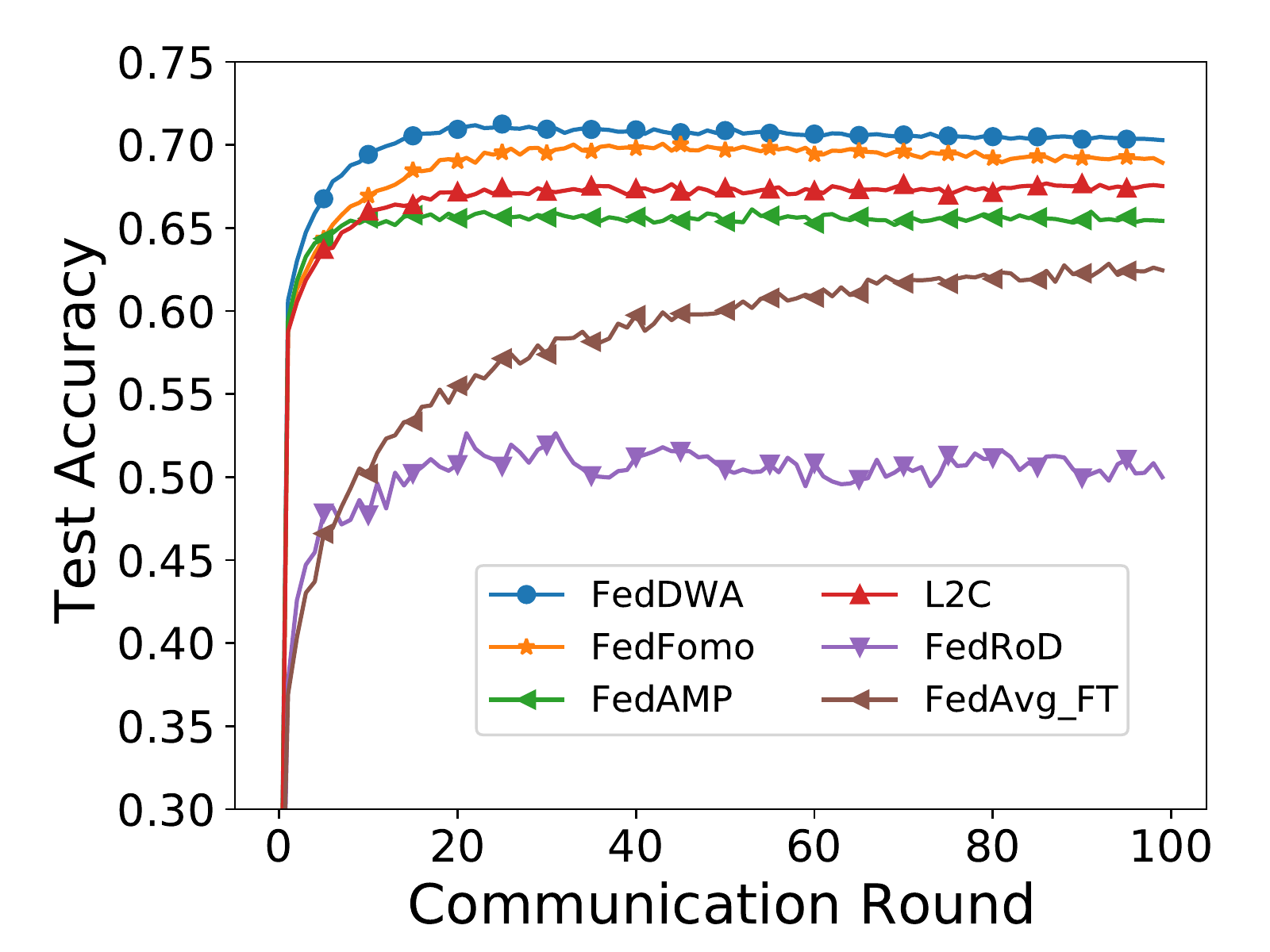}% Reduce the figure size so that it is slightly narrower than the column. Don't use precise values for figure width.This setup will avoid overfull boxes.
\label{appendix: appendix_cinic10_noniid10_curve}
}
\caption{Test accuracy over communication rounds under the practical heterogeneous setting 1 with 20 clients.}
\label{appendix: appendix_noniid10_curve}
\end{figure*}

\subsection{Performance of Guidance Model}
\begin{figure*}[t]
\centering
\includegraphics[width=1.5\columnwidth]{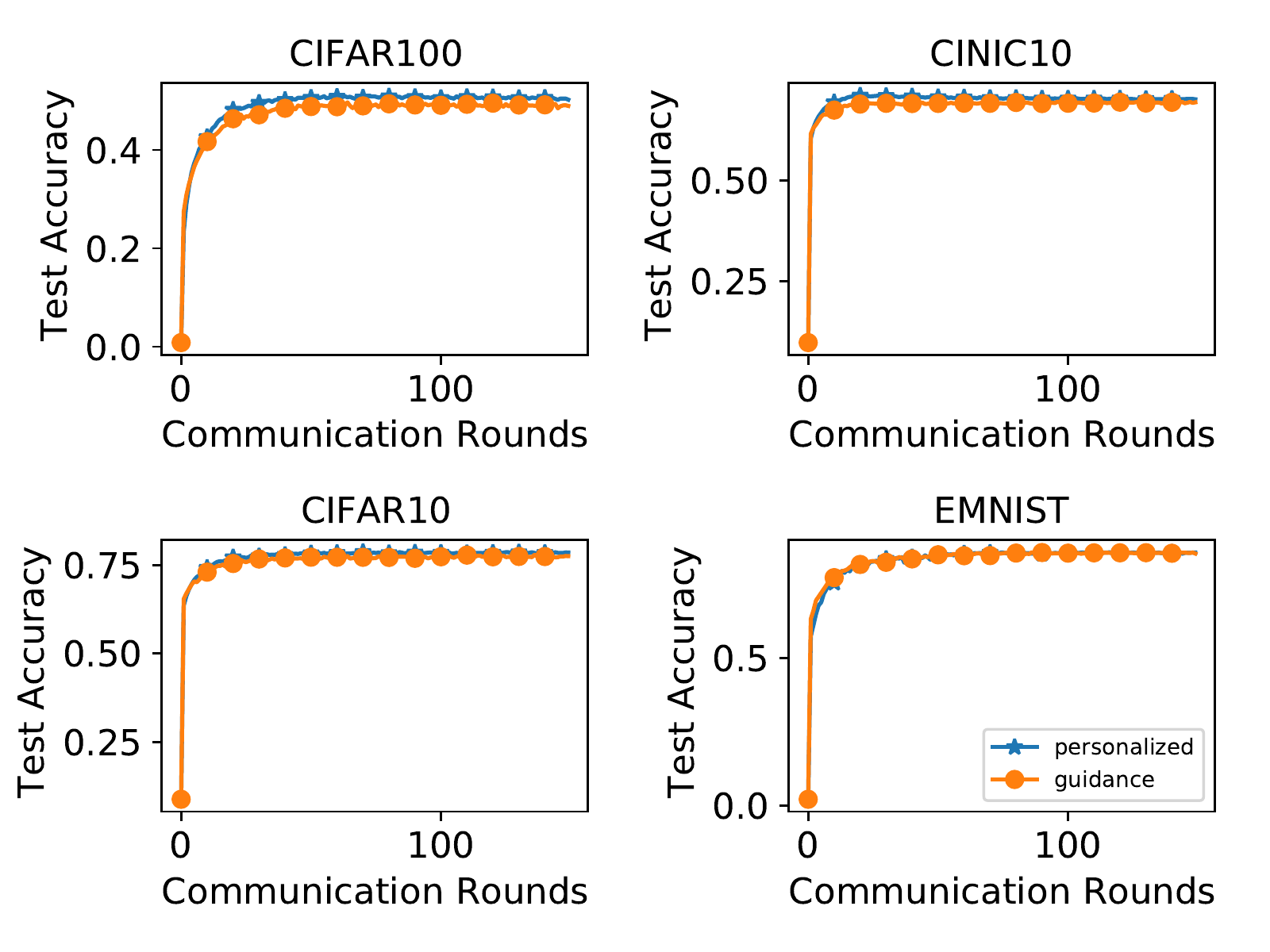}% Reduce the figure size so that it is slightly narrower than the column. Don't use precise values for figure width.This setup will avoid overfull boxes.
\caption{Test accuracy over communication rounds under the practical heterogeneous setting 1 with 20 clients.}
\label{appendix: guidance_curve_noniid10}
\end{figure*}

\begin{figure*}[t]
\centering
\includegraphics[width=1.5\columnwidth]{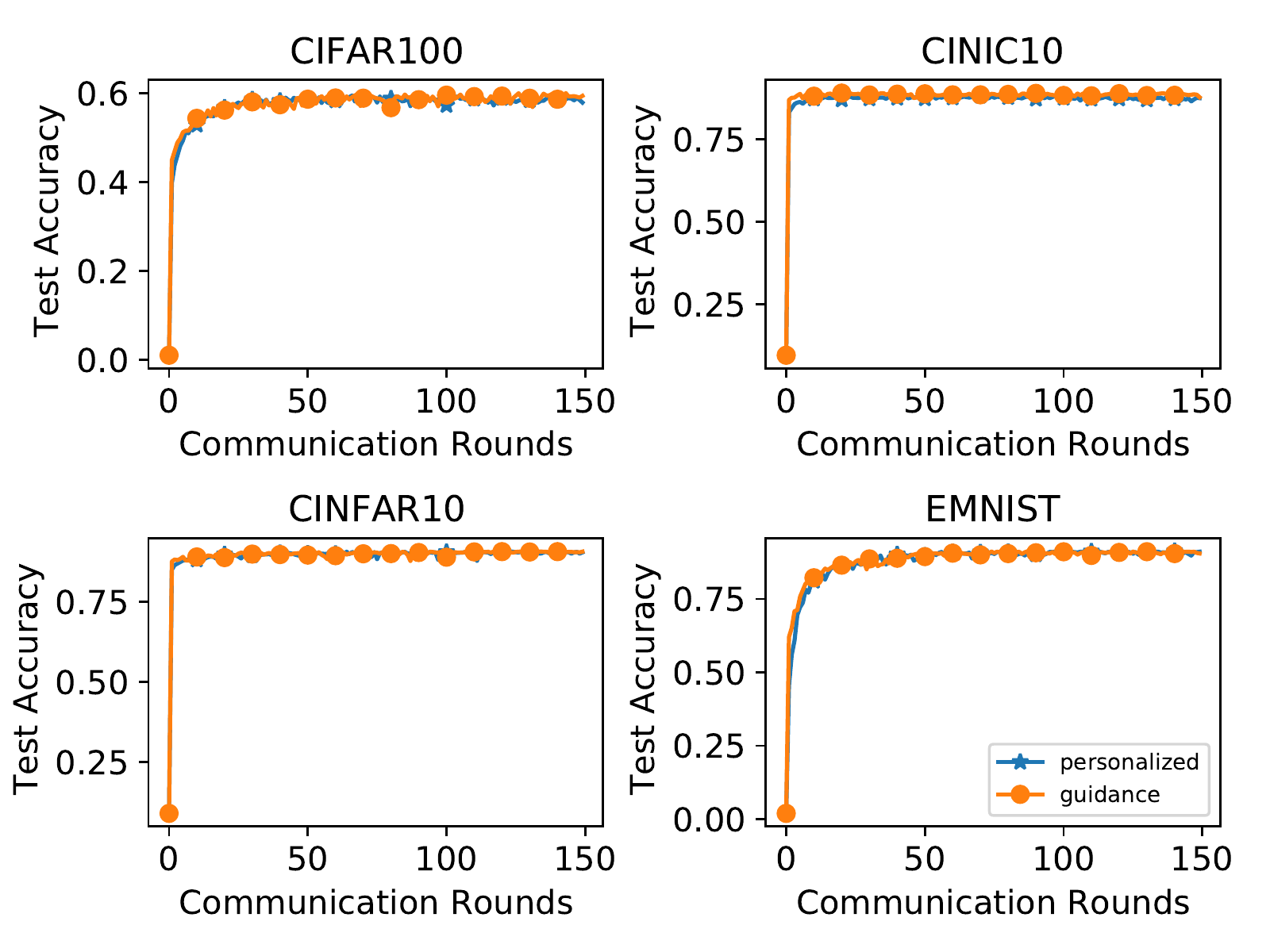}% Reduce the figure size so that it is slightly narrower than the column. Don't use precise values for figure width.This setup will avoid overfull boxes.
\caption{Test accuracy over communication rounds under the practical heterogeneous setting 2 with 20 clients.}
\label{appendix: guidance_curve_noniid9}
\end{figure*}

As we have  mentioned in the main body of the paper, guidance model $\hat{w}_{i}^{\star}$ represents the data distribution of client $i$. From this point of view, one-step ahead adaptation is a reasonable approximation. $\hat{w}_{i}^{\star}$ takes one-step ahead of time such that it can instruct client $i$ to identify other clients which should be assigned with higher weights for model aggregation. Through experiments, we will show that $\hat{w}_{i}^{\star}$ is effective because its performance is similar to that of the personalized model $w_{i}^{t}$. Table \ref{appendix: tab_comparison1} lists the average test accuracy of guidance model and personalized model after training 150 rounds, under the practical heterogeneous setting 1 and the practical heterogeneous setting 2 ($\alpha=0.1$), respectively. It can be seen that the test accuracy of both is almost the same in the practical heterogeneous setting 1. In addition, we find that the performance of the personalized model is slightly better than that of the guidance model in the practical heterogeneous setting 2, which means that when the data distribution between clients has a clustering structure, our algorithm can exactly capture this similarity to benefit clients. The training curves are presented  in Figure \ref{appendix: guidance_curve_noniid10} and Figure \ref{appendix: guidance_curve_noniid9}.
% and we can get the similar conclusion  from our optimization problem (Eq.~\eqref{problem2}), since the meaning of Eq.~\eqref{problem2} is looking for personalized weights that minimize the distance between guidance model and personalized model.

\begin{table}[h]
\centering
\begin{tabular}{cccc}
\hline
Dataset & Practical Setting & \multicolumn{1}{c}{\begin{tabular}[c]{@{}c@{}}Guidance \\ model\end{tabular}} & \multicolumn{1}{c}{\begin{tabular}[c]{@{}c@{}}Personalized \\ model\end{tabular}} \\ \hline
EMNIST & Setting 1 & 85.96 & 86.00 \\ \cline{2-4} 
 & Setting 2 & 91.26 & 91.26 \\ \hline
CIFAR10 & Setting 1 & 77.86 & 78.67 \\ \cline{2-4} 
 & Setting 2 & 90.66 & 90.62 \\ \hline
CIFAR100 & Setting 1 & 49.66 & 51.10 \\ \cline{2-4} 
 & Setting 2 & 59.44 & 59.54 \\ \hline
CINIC10 & Setting 1 & 69.60 & 71.17 \\ \cline{2-4} 
 & Setting 2 & 87.74 & 87.47 \\ \hline
\end{tabular}
\caption{The best test accuracy (\%) over four different datasets under two practical heterogeneous setting with 20 clients.}
\label{appendix: tab_comparison1}
\end{table}

\subsection{Curve of Test Accuracy During Training}
\label{appendix:comparison_results}
Figure \ref{appendix: appendix_noniid8_curve} and Figure \ref{appendix: appendix_noniid10_curve} present the evolution of average test accuracy over global communication rounds for partial experiments shown in Table \ref{tab1}. From which, it can be seen that our method has a significant performance improvement compared with other methods, except in the pathological heterogeneous setting with CINIC10 dataset. 
% {\bf YP: why you need to present the figures since you already have tables? What insights the table cannot tell us such that you have to show figures?
% It is very unsually to introduce figures presented earlier. 
% the figure sequence should corresponds to discussion sequence. }

\subsection{Communication Cost}
\label{appendix:communication_cost}
\begin{table}[h]
\centering
\begin{threeparttable}
\begin{tabular}{lccc}
\hline
 & \multicolumn{1}{c}{Communication} & \multicolumn{1}{c}{CIFAR100} & \multicolumn{1}{c}{TINY} \\ \hline
FedAvg & $2\times \Sigma$ & 9.54 & 53.56 \\
Others\tnote{*}  & $2\times \Sigma$ & 9.54 & 53.56 \\ \hline
FedFomo & $(1+M)\times \Sigma$ & 28.62 & 160.68 \\
L2C & $(1+N)\times \Sigma$ & 100.17 & 562.38 \\
Ours & $3\times \Sigma$ & 14.31 & 80.34 \\ \hline
\end{tabular}
\begin{tablenotes}
\item$^*$ \small{Others includes FedAvgM, FedRoD, FedAvg\_FT, FedProx, per-FedAvg, pFedMe, SFL and  ClusterFL}.
\end{tablenotes}
\end{threeparttable}
\caption{The amount of the communication traffic (MB) incurred by each client per round for different algorithms. $\Sigma$ is the size of the model. $M (M\ge 1)$ and $N$ are the number of models downloadad by each client. We set $M=5$ and $N=20$, as in the orginal papers. }
\label{tab3}
\end{table}
The amount of data (including upload and download) each client needs to transmit  per communication round by using different methods is compared in Table \ref{tab3}. 
% Here $M$ is a hyper-parameter of FedFomo and means that each client needs to download $M$ models from other clients. In our experiment, we set $M=5$ according to the original paper.  $N$ means the number of total clients, and we set $N=20$. 
In comparison with FedFomo and L2C, it is apparent that our method can sheer shrink communication traffic for detecting client similarity. 
Although FedFomo, L2C and our method incur more communication traffic than that of FedAvg and others, a notable advantage of FedFomo, L2C and our method is that they can explicitly measure and explain the similarity between clients, which have be explored in Figure \ref{similarity}.

\subsection{Computationbal Cost}
\label{appendix:computational_cost}
Suppose that there are $N$ clients participating in training and the number of model parameters is $d$, the computation complexity of FedDWA (ours), FedAMP and L2C are all $ \mathcal{O}(N^2d)$ in the server. For FedFomo, the extra computation is offloaded on clients, and thus its complexity in the server is $\mathcal{O}(Nd)$. We also test the total FLOPs for each communication round using CIFAR10 dataset, and the results can be found in Table \ref{appendix: table_computation}. In FedDWA, the magnitude of FLOPs needed to calculate the similarity (Eq.(\ref{solution3})) is $10^{8}$ while it is $10^{11}$  for model training, indicating that computation load is mainly generated by model training. 

\begin{table}[h]
\centering
\begin{tabular}{lc}
\hline
Methods      & FLOPs              \\ \hline
FedAvg       & $2.5\times10^{11}$ \\
FedFomo      & $4.6\times10^{11}$                   \\
FedAMP       & $2.5\times10^{11}$                   \\
L2C          & $1.2\times10^{12}$                   \\
FedDWA(ours) & $5.1\times10^{11}$                   \\ \hline
\end{tabular}
\caption{The amount of the computational cost incurred by each client per round for different algorithms.}
\label{appendix: table_computation}
\end{table}

\subsection{More Discussion About Guidance model}
\label{appendix:selection_of_guidance_model}

 \subsubsection{What is guidance model.}
\emph{A guidance model can facilitate the training of personalized models by enabling collaborations between similar clients, %, resulting in better performance than local training only.
} However, it's difficult to directly define the optimal guidance model since it should be the optimal personalized model, i.e., the objective of PFL. We have tried different choices of the guidance model (see next section), and  one-step-ahead adaptation is the best one among our trails.  The guidance model can be intuitively interpreted as follows. At the beginning of  round $t+1$,  client $i$  downloads the global model $w_{t}$, which actually guides the learning of client $i$ as $$w_{i}^{t+1} \gets  w_{i}^{t} + \sum_{j=1}^{N}p_{i,j}\cdot (w_{j}^{t} - w_{i}^{t}),$$ where $ w_{t}  = \sum_{j=1}^N p_{i,j} w_j^t$  is the aggregation of models in round $t$ and $\sum_{j}p_{i,j}=1$ is aggregation weights. 
How to set $p_{i,j}$ is important to achieve PFL. Traditional FL set $p_{i,j}=\frac{1}{N}$, which is an unbiased estimate of global model and does not take into account the unique target of each client. 
%This then updates client $i$'s current local model in directions specified by the weight $\bf{p}_{i}$ and $\{w_{j}^{t}\}$. However, traditional FL set $p_{i,j}=\frac{1}{N}$, which is an unbiased estimate of global model and does not take into account the unique target of each client. 
FedFomo and L2C want to find the optimal aggregation weights %$\bf{p}_{i}$ $=[p_{i,1},...,p_{i,N}]$ 
to optimize personalized models on individual clients. %objective of client $i$. 
However, as described in the paper, their methods will incur huge communication overhead with the risk of privacy leakage. Our work improves these defects by introducing a guidance model 
to guide the setting of $p_{i,j}$ by constructing the optimization problem (in Eq.~(9)). Through Eq.~(9), we subtly offload  the computation of personalized aggregation weights to the server to  reduce the communication cost. Besides, our guidance model can  characterize client similarity in an analytical way rather than an empirical search via the validation dataset. 

\subsubsection{How to select guidance model.}
As we have point out in the text, we can use the last iteration model $\hat{w}_{i}^{t-1}$, the current model $w_{i}^{t}$ or the local one-step ahead adaptation model $\hat{w}_{i}^{t}-\eta_{i}^{t-1}\nabla f_{i}(\hat{w}_{i}^{t})$ for the guidance model. We have conducted  experiments to select the guidance model for both Pathological Setting (CIFAR10) and Practical Setting 1 (CIFAR10-V2), and the results are shown in Table \ref{appendix:tab_selection_guidance_model}.  Here $w$ represents the model after one-step ahead adaptation and $w_{1}$ represents the model in the last iteration, and we find that the performance of using the current model $ w_{i}^{t}$ is inferior to the others, so it's not shown here. In Table \ref{appendix:tab_selection_guidance_model}, it can be seen that  using $w$ and $w_{1}$ achieves similar results. However, if we use $w_{1}$, we have to store $w_{1}$ with two possible cases:
i) Store $w_{1}$ in clients who need to upload two models to the server;
ii) Store $w_{1}$ in the server, so that each client only needs to upload one model (in this case, the amount of uplink traffic and download traffic is the same as that of FedAvg). For both cases, we need extra memory for storing the historical model $w_{1}$. Besides, if we use the last iteration model $w_1$, then Eq.(\ref{solution3}) will rely on the historical information, which may lead to the cold-start problem when there are newly-joined clients. Instead, if we use $w$, we can save the memory overhead to achieve a similar performance and at this time, since Eq.(\ref{solution3}) doesn't rely on historical information, FedDWA is not sensitive to the cold-start problem. The experimental results in Table \ref{tab2} also confirm this point. Therefore, we finally choose $w$.

\begin{table}[h]
\centering
\begin{tabular}{lccc}
\hline
 & CIFAR10 & CIFAR10-V2 & Communication \\ \hline
$w$ & 92.97\% & 78.56\% & $3\times \Sigma$ \\
$w_{1}$ & 92.22\% & 78.99\% & $2\times \Sigma$ or $3\times\Sigma$ \\ \hline
\end{tabular}
\caption{Final test accuracy and communication overhead per round.}
\label{appendix:tab_selection_guidance_model}
\end{table}

% Please add the following required packages to your document preamble:
% \usepackage{multirow}

\end{document}